\documentclass[pdflatex,sn-nature]{sn-jnl} 

\usepackage{graphicx}%
\usepackage{multirow}%
\usepackage{amsmath,amssymb,amsfonts}%
\usepackage{amsthm}%
\usepackage{newtxtext,newtxmath}
\usepackage[title]{appendix}%
\usepackage{xcolor}%
\usepackage{textcomp}%
\usepackage{manyfoot}%
\usepackage{booktabs}%
\usepackage[linesnumbered,ruled,vlined]{algorithm2e}
\usepackage{algorithmicx}%
\usepackage{algpseudocode}%
\usepackage{listings}%
\usepackage{lineno} 

\theoremstyle{thmstyleone}%
\theoremstyle{thmstyletwo}%
\theoremstyle{thmstylethree}%

\begin{document}

\title[Article Title]{A Neural Network Alternative to Tree-based Models}


\author*[1,2]{\fnm{Raieli} \sur{Salvatore}}\email{salvatore.raieli2@gmail.com}
\equalcont{These authors contributed equally to this work.}

\author[1]{\fnm{Jeanray} \sur{Nathalie}}
\author[1]{\fnm{Gerart} \sur{Stéphane}}
\author[1]{\fnm{Vachenc} \sur{Sebastien}} 
\author*[2]{\fnm{Altahhan} \sur{Abdulrahman}}
\email{a.altahhan@leeds.ac.uk.}
\equalcont{These authors contributed equally to this work.}

\affil[1]{ \orgname{Oncodesign Precision Medicine}, \orgaddress{\street{18 rue Jean Mazen}, \city{City}, \postcode{21000},  \country{France}}}

\affil[2]{ \orgdiv{School of Computer Science}, \orgname{University of Leeds}, \orgaddress{ \country{UK}}}

\abstract{Tabular datasets are ubiquitous in biological research, where AI-driven analysis is increasingly adopted. While interpretable tree-based models currently dominate, they lack the scalability of artificial neural networks (ANNs), which excel in complex non-tabular tasks due to their retrainability and integration capabilities. However, ANNs often underperform compared to tree-based methods on tabular data, both in accuracy and interpretability. This work introduces sTabNet, a meta-generative framework that dynamically generates ANN architectures specifically optimised for tabular datasets. sTabNet enforces sparsity using an unsupervised approach that focuses on features rather than datapoints, enabling efficient capture of important feature relationships. Evaluated across diverse biological tasks (RNA-Seq, single-cell, survival analysis), sTabNet outperforms state-of-the-art tree-based models like XGBoost and provides a scalable foundational model that can be fine-tuned for both in-domain and out-of-domain tasks. Furthermore, sTabNet offers built-in interpretability via attention mechanism, providing direct insights into feature relevance and outperforming post-hoc explainability methods like SHAP.}




\maketitle

\section{Introduction}
Although tabular data are widespread, they have been left behind by recent advances in artificial intelligence \cite{Borisov_2022}. Recent literature suggests that tree-based models (such as Random Forest or XGBoost \cite{pgb2023}) outperform neural networks for tabular data \cite{grinsztajn2022treebased}. A natural advantage of a neural network model is the ability to perform other tasks, with minimal training, via transfer learning or fine-tuning.

Despite a wide interest in neural networks for tabular data, up to date, tree-based models are considered state-of-the-art for medium tabular datasets. Over the years, several large models have been proposed without a consistent gain in performance over tree-based models \cite{rubachev2022revisiting}.  In addition, the lack of a tabular benchmark has made it difficult to compare different methods \cite{grinsztajn2022treebased}. As shown in \cite{grinsztajn2022treebased}, while large capacity models are computationally intensive, they do not have superior performance. In particular, they often bear a performance gap compared to tree-based models. Thus, simpler, high-performing, and more efficient neural network models are needed.
    
For neural network models to constitute a viable alternative to the tabular domain, they need to perform competitively with tree-based models, provide meaningful feature importance and be interpretable, preferentially without the need for post-hoc methods. This is especially true for biomedical tabular data (such as omics data) where interpretability and model decision transparency lead to a direct impact on medical applications: new drug targets, biomarkers, clinical decisions, and risk factors \cite{novakovsky2023obtaining}, \cite{watson2019clinical}. 
     
Typically, current proposed neural network models to tackle tabular domain have dense connectivity leading to a relatively higher number of parameters, which leads to learning a complex function but at the risk of overfitting. In other words, dense networks can learn spurious correlations and memorize exceptions (outliers, noise etc.), leading to poor performance even for simpler tabular tasks \cite{bayat2024pitfallsmemorizationmemorizationhurts, Roberts_2021}. Nevertheless, a strong interest remains in developing neural network models for such data \cite{Borisov_2022} for better integration with medium and large neural network models, as well as preserving flexibility and versatility \cite{somvanshi2024surveydeeptabularlearning}. 
    
Widely used, medium-size, versatile models such as Resnet, Mobilenet, and VGG have not been designed with tabular data in mind. Likewise, large-size foundational models such as GPT and BERT are overly bloated and not designed to tackle tabular problems. Both of these kinds of models have high complexities that necessitate training beyond the typical size of tabular data, making them hard to tailor and specialize for such domains \cite{bommasani2022opportunities}. A sleek, simple, versatile but foundational neural network architecture is required to address the needs of tabular domains for the long run.
    
As noted in the lottery ticket hypothesis, most of the weight in a neural network could be pruned  \cite{frankle2019lottery}. However, while sparsity is a desirable model attribute, it is hard to achieve before training, and consequently, post-training methods are used (such as pruning) \cite{liu2023lessons}.  Nevertheless, sTabNetworks have been leveraged for tailored applications with genomics data \cite{wysocka2023systematic} where the sparsity-introduction process is driven by prior knowledge, but they still lack interpretability. While this is a particular case, it has not been extended to other tabular datasets, and prior knowledge does not lend itself to generalisation. 
    
Interpretability, on the other hand, is required in many applications (such as biology, banking, and insurance) where algorithm predictions directly impact human quality of life \cite{DeloitteXAI}. Given that governmental institutions are also considering regulations for artificial intelligence applications, interpretability for tabular models is an important but unmet need \cite{DeloitteXAI}.
           
In this paper, we introduce a foundational model that attempts to address the above-mentioned shortcomings of neural networks by proposing a general-purpose algorithm to build and train a sparse interpretable neural network model that can be tailored for the tabular problem in hand with little to no training or fin-tuning. 

\section{Background}

In \cite{wysocka2023systematic}, biologically constrained neural networks are introduced in the biomedical domain as feed-forward neural networks (FFNN), where sparsity is introduced before training. To achieve sparsity, the authors utilize the knowledge contained in an external database to define the model architecture.  \cite{Elmarakeby_2020} used the Reactome database to define the interconnectivity between the features(genes) by utilizing the dependencies between them. The data can be seen as a matrix of overlapping clusters of multi-features (called pathways in biology) (Fig. 1A). The feature interactions in the modified linear layer is controlled via a binary matrix. Thus, if there is biological knowledge about the interaction of two features (belonging to the same cluster), a value of 1 is assigned in the binary matrix; otherwise, the value is 0. This matrix is masked/multiplied element-wise with the FFNN weight matrix (Fig. 1B). 
    
The advantage of this method is that it incorporates domain knowledge in the model, and enhances the transparency of the model \cite{wysocka2023systematic}. However, post-hoc explainability methods (such as SHAP) are still required to interpret the model \cite{wysocka2023systematic}. Thus, it remains to define interpretation mechanisms that allow for the identification of the important features. While this approach can be applied to biomedical datasets, a similar sTabNet cannot be determined for datasets where external domain knowledge is lacking. 
       
Inspired by the core competencies provided by attention mechanisms, we define a simple but interpretable mechanism for tabular data. We show that sTabNet with this mechanism outperforms tree-based models for biomedical tabular datasets. Moreover, we show that our approach is better at identifying the most relevant features in the dataset than post-hoc methods and supports in-domain and out-of-domain adaptation. Finally, we defined a simple algorithm to generalize the benefits of such sTabNet and attention mechanisms to any tabular dataset, dubbed sTabAlgo. Our sTabNet stemming from sTabAlgo is shown to be competitive with tree-based models even without hyperparameter search.   

\section{sTabNet: A sparse model for tabular data}
\begin{figure}[!ht]
        \centering
         \includegraphics[width=12cm, height=12cm]{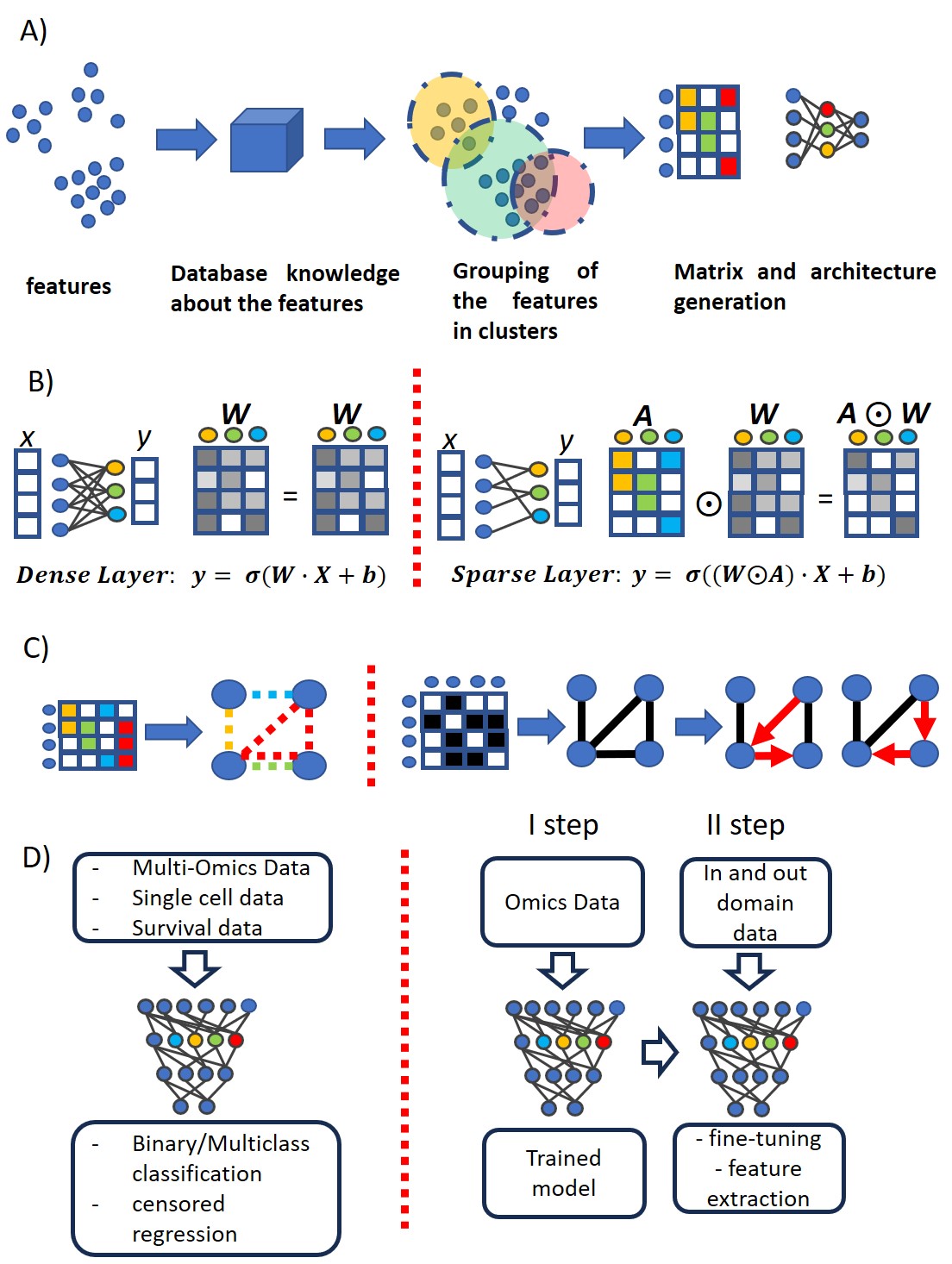}
        \caption{ A sparse and interpretable neural network. \textbf{A}. sTabNet Architecture: Features can be grouped according to prior knowledge or by using unsupervised learning (clustering or random walk) to build a matrix A where rows are features and columns are clusters (neurons). In this sparse model, a feature is connected to a neuron (which represents a cluster) only if it is a member of the cluster. \textbf{B}. sTabNet Sparsity:The representation and the definition of the classical dense layer (left) and of the proposed sparse layer (right). The sparse layer is identical to the dense layer except for the Hadamard product between the weight matrix \textbf{W} and the matrix \textbf{A}.  \textbf{C}. sTabNet Grouping: (left) The matrix \textbf{A} can be intended as a compressed view of an adjacency matrix of the feature graph. The neuron can also be defined as a random walk in the feature graph, thus learning a local approximation of the neighborhood of a feature. Alternatively, one can use clustering (not shown in the figure). (right) Unrolling of the process in the left: When information about features in a dataset is not present, we calculated the cosine similarity matrix of the features. We assigned an edge between two features if their similarity is higher than 0.5. We performed random walks on the obtained graph and used the obtained random walks to build the sparse matrix in the modified layer. \textbf{D}. sTAbNet as a Tabular Foundational Model: A scheme of sTabNet used for different tasks and data types. The same architecture can be used for common and challenging biological tasks (binary/multiclass classification, censored regression) and complex data (RNAseq, single cell, and multi-omics data). . sTabNet has been tested with real-world datasets for all these tasks. On the left, we are showing that the model can be trained on a dataset, and then the trained model can be used for other datasets and tasks through fine-tuning or feature extraction. }
        \label{fig:1}
    \end{figure}

We aim to build a high-performant, efficient, interpretable neural network model for tabular data. High performance and efficiency can be achieved via sparsity, which we discuss first, and then we discuss achieving interpretability. 
We impose sparsity a priori by constraining the connectivity of the neurons in the model architecture to restrict the number of trainable parameters in the model and reduce the tuning required to reconcile the contribution of these weights to different types of data points. This, in turn, allows the model to commit different parts to different data varieties more effectively. This curbed connectivity is achieved by grouping the features according to their innate properties and allowing them to be connected only to the neurons in subsequent layers representing their common innate properties.  Figure \ref{fig:1} shows this process.

We can uncover the innate properties of features either by employing unsupervised learning methods, such as clustering or random walk, or by exploiting prior domain knowledge when it is available. In either case, we are grouping/clustering the features, not the data points. In other words, the intrinsic groups/clusters corresponding to neurons in the model are specified by the features' similarity, not the data points' similarity. A feature point consists of a vector; its components are the feature readings for all the data points in the dataset. 
Figure \ref{fig:S8} shows the comparison between using unsupervised learning and domain knowledge to construct the sTabNet model, where we show that in general model built using unsupervised learning performs comparable to the models built using domain knowledge.

If $X^{m\times n}$, where $m$ is the number of data points and $n$ is the number of features, $A$ is a design matrix capturing the dataset. Then our grouping algorithms will operate on the $n$ dimension to create a set of clusters $K<n$  grouping the features. Note that we usually apply a clustering algorithm to the $m$ dimension to group the data points, not the features.

This grouping process dictates the number of neurons in the model, specifying the connectivities between the input layer and the first hidden layer. Features connected to the neurons are only those that belong to the group/cluster represented by this neuron. 

When prior information about the features is available, it can be exploited to construct a matrix to control the structure of the neural network (Fig. 1A). For example, for RNA expression datasets, one can use the information contained in a pathway database to control which features can be grouped together. This binary matrix is then used in a masking operation to control the sparsity of the neural network (Fig. 1B). 

When one does not have specific domain knowledge about the dataset, we can construct this matrix with a simple grouping or clustering process. Considering the similarity between features in the dataset (cosine similarity or another measure of similarity), one can either apply one of the well-known clustering algorithms or construct a feature graph and use random walks to build the binary matrix. In short, explore the various global or local feature interactions depending on the hyperparameters (Fig. 1C-D). 

\section{sTabNet interpretability for tabular data}
Next, we defined simple attention mechanisms adapted for tabular data based on feature importance. This avoids the need to use post-hoc methods to calculate feature importance. Feature importance is determined during training, not after the model has already been trained. We tested our sTabNet to verify that the attention mechanisms captured the importance of features for tabular datasets (exploiting both simulated and real datasets). We also tested sTabNeton real and complex datasets, such as multi-omics, single cells, multiclass classification, and survival regression, comparing them with decision trees (Fig. 1E).

In the next section, we provide a detailed walkthrough of both the sparse matrix construction (both in the presence of feature information and in an unsupervised manner) and the attention mechanism.

\subsection{Interpretability, attention and feature importance}

As highlighted by \cite{Borisov_2022}, interpretability measurements are complicated because we do not have a dataset where the importance of the features is known in advance (a ground truth). Therefore, to test the effectiveness of our interpretability approach, we used a set of synthetic datasets with various difficulties. In this evaluation framework, the ground truth of feature importance and the level of complexity can be controlled a priori via a redundancy and separation coefficient, respectively. Figure 2 shows this set of experiments. 

First, from a classification perspective, we observed an expected linear decrease of XGBoost accuracy with the increase in the difficulty of the multi-classification task (Fig. 2A) where the classes are harder to separate. When the classes are poorly separated (separation coefficient 0.1), we observed a reduction in the weight assigned to important features making it harder to separate them from noisy features (Fig. 2B) and the model where almost doing a random guessing. The SHAP value follows the same pattern for the feature importance, not allowing better identification of important features (Fig. 2C). Thus, SHAP explanations in a sense are the expression of the prediction abilities of XGBoost, they are dependent on its accuracy, but they do not improve the ability of the model in capturing the feature importance (and to separate informative and noisy features).

We incorporate a tabular attention mechanism (defined in the method section) to automatically define the importance of the input features. While we also observed decreased sTabNet performance with increasing difficulty in the multiclassification task, the reduction is less dramatic (Fig. 2D and S1 A-C). Moreover, the separation between important features and noisy features is more pronounced (Fig. 2E). Interestingly,  when the classification task is more complex, sTabNet needs longer training to reach better performance and better separation (increasing the number of epochs) (S2 A-B).

We conducted a feature ablation study to examine the relationship between feature importance and attention. Our hypothesis is to use attention weights as aliases for features' importance. The most relevant first (MoRF) analysis shows a sharp drop when a feature with high attention weight is removed, demonstrating that attention mechanisms identify discriminative/important features (Fig. 2F). 

The least relevant first (LeRF) analysis shows that attention scores are identifying noisy features (Fig. 2G). When the number of noisy features is increased, the associated XGBoost features' importance decreases (Fig. 2H). However, the attention feature importance does not change for our proposed network sTabNet, allowing a separation between important features and noise (Fig. 2I). Finally, we also compared with other commonly used feature importance methods and noticed a general concordance (supplementary table 1). 
These results show that the \textit{attention weight can be considered a feature importance score}. It is a better discriminator between important features and noise than XGBoost feature importance and SHAP value (Fig. 2C). In addition, they demonstrate that the attention score is more stable in separating significant versus noisy features.

\begin{figure}[h!]
        \centering
         \includegraphics[width=13cm, height=14cm]{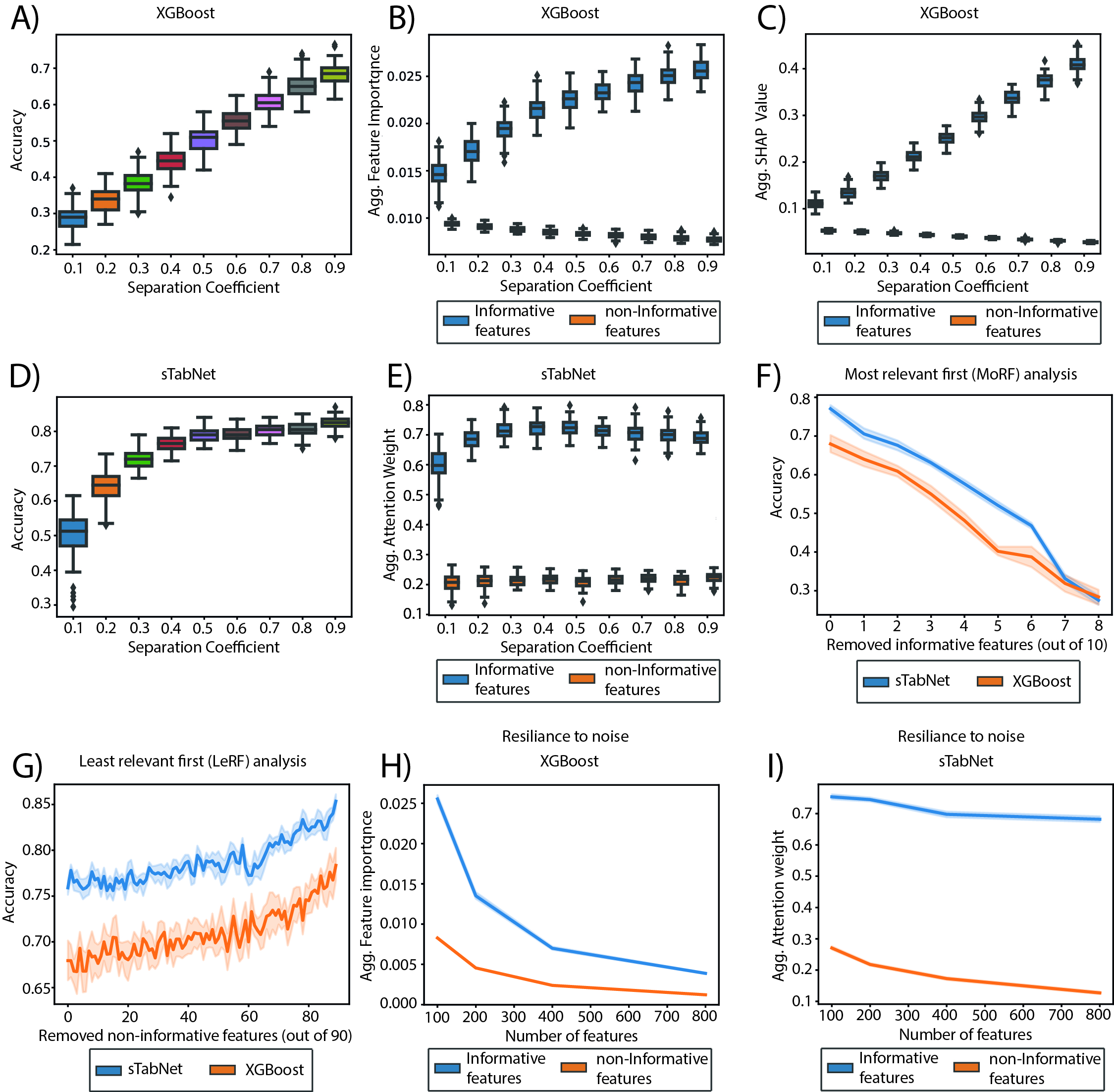}
        \caption{Attention mechanisms are a measure of feature importance. \textbf{A-F}. Each boxplot represents 100-fold hold-out validation; a lower coefficient represents a harder multiclassification task. \textbf{A}. Multi-classification accuracy in XGBoost with an increase in separation difficulty.  \textbf{B}.  Separation between the average importance weight (XGBoost's feature importance) assigned to real informative and non-informative features. \textbf{C}. Separation between the average importance weight (SHAP value) assigned to informative and non-informative features. \textbf{D}.  Multi-classification accuracy in sTabNet with an increase in separation difficulty. \textbf{E}. Separation between the average importance weight (feature attention weight) assigned to real informative and non-informative features. \textbf{F-G}. Accuracy is plotted for each different model (XGBoost and sTabNet). The shade represents the standard deviation (10 different models for each removed feature).  \textbf{F}. MoRF analysis. \textbf{G}. LeRF analysis. \textbf{H-I}. The shade represents the standard deviation (10 different models trained). \textbf{H}. Feature importance for XGBoost when increasing the number of non-informative features. \textbf{I}.  Feature importance for the sTabNet with an increase in the number of non-informative features.}
        \label{fig:2}
    \end{figure}

\section{sTabNet on complex real-world datasets}
We decided to test our model with complex multi-omics real-world data, which presents different challenges for machine learning models \cite{reel2021using, lopez2019challenges, vahabi2022unsupervised, attwaters2023bridging}. Our sTabNetwork is built using METABRIC multi-omics data as described above. Since there is no consensus in the literature on the best activation function for biologically constrained networks \cite{Elmarakeby_2020, hao2018pasnet}, we tested different ones during training. However, we did not observe any particular effect due to the choice of the activation function (Fig. S3A). 

More importantly, we also noticed that when we fit a feed-forward neural network it only predicts the majority class with this dataset. Thus, we did not include ulterior experiments with a fully connected neural network in this study due to this triviality. Neither the convolutional neural network is showing satisfying performances (Fig. 3A). 

Instead, the sTabNet shows results comparable to those of XGBoost (Fig. 3A). Moreover, prepending an attention layer shows a slight increase in performance (Fig. S3B and supplementary Table 2). Since we do not have the ground truth for feature importance for this dataset, we have selected the 100 features with the highest attention weights, and we conducted a gene-set enrichment \cite{subramanian2005gene} for diseases on this feature subset. The results show that the features with the highest attention weights are attributed to cancer-related genes (Fig. 3E-F and supplementary table 3), which is consistent with the results from the synthetic dataset.

\subsection{sTabNet and transfer learning}
Since an advantage of neural networks is the possibility to fine-tune a model on a different dataset, we tested if a model trained on METABRIC, which has a multiclass objective, could be fine-tuned for a binary classification dataset (TCGA-BRCA). The fine-tuned model demonstrated the ability to adapt to a different task and a different in-domain dataset (in-domain adaptation) (Fig. 3B). Furthermore, the frozen model can extract meaningful features that can be used to train a linear classifier (Fig. 3B). We also noticed out-of-domain adaptation using TCGA-LUAD. Both fine-tuning model and feature extraction showed good results (Fig. 3B).

\subsection{sTabNet on single-cell technology}
Single-cell technology has had a transformative impact on our understanding of development and the role of a cell in both physiological and disease environments. At the same time, these types of datasets have different computational challenges, mainly due to having more features than data points, and there is currently no consensus on how to tackle this problem \cite{stuart2019integrative, baysoy2023technological, lim2023transitioning}. We used single-cell data from the Tumor Immune Single-cell Hub 2 (TISCH2) of breast cancer (GSE161529). We used sTabNet for binary classification (tumour/normal cell) and multi-classification tasks (cellular type prediction). We did ten cross-fold validations for each model, averaged obtained values, and reported the accuracy. After freezing the model, we also extracted the data representation and performed data dimensional reduction with UMAP to plot it. The sTabNet shows impressive performance on the more complex single-cell data in both binary and multiclass classification settings (Fig. 3D and Supplemental Table 4). Moreover, the model-learned representation of the data allows for the separation of the classes (Fig. S4), which corresponds with the assumption that reduced connectivity helps commit certain model parts to a certain variety of data. This shows that the model is learning meaningful features of biological datasets that could be used for follow-up tasks. More importantly, since genomic datasets suffer from the curse of dimensionality \cite{chattopadhyay2019gene}, sTabNet can be used efficiently for \textit{feature selection} to reduce the number of features and the model complexity in an analysis pipeline (Fig. S.5).

\subsection{sTabNet and survival analysis}
Survival analysis is a domain of machine learning that analyzes the dependence of time on the occurrence of an event (death, failure, and so on) and its relationship to features \cite{george2014survival, wang2019machine}. Although it can be considered a subcase of regression, classical techniques fail to capture its complexity \cite{sarkar2017analysis}. Given the importance and the challenge presented by the survival dataset for biological studies, several models have been developed to address this task, such as the Cox proportional hazards, but these models fail in nonlinear situations and when they deal with large datasets \cite{cox1972regression, mariani1997prognostic}. 
For this reason, we compared the sTabNet with the latest models from the Scikit-survival library, which integrates several methods based on traditional machine learning \cite{sciki_surv}. We show that a sTabNet clearly outperforms different machine learning algorithms for survival analysis (ensemble tree-based and support vector machine-based) (Fig. 3D). In addition, it can extract the most important predictors of survival (Fig. S.6).

The results show that sTabNet with attention mechanisms performs better than tree-based models for genomic datasets. They are interpretable, and since they learn meaningful latent features, they permit in-domain and out-of-domain adaptation in a foundational manner. Moreover, they can be used for complex biology tasks such as single-cell and multi-omics classification and survival regression as seen in Figure 3.

\begin{figure}[h!]
        \centering
         \includegraphics[width=14cm, height=15cm]{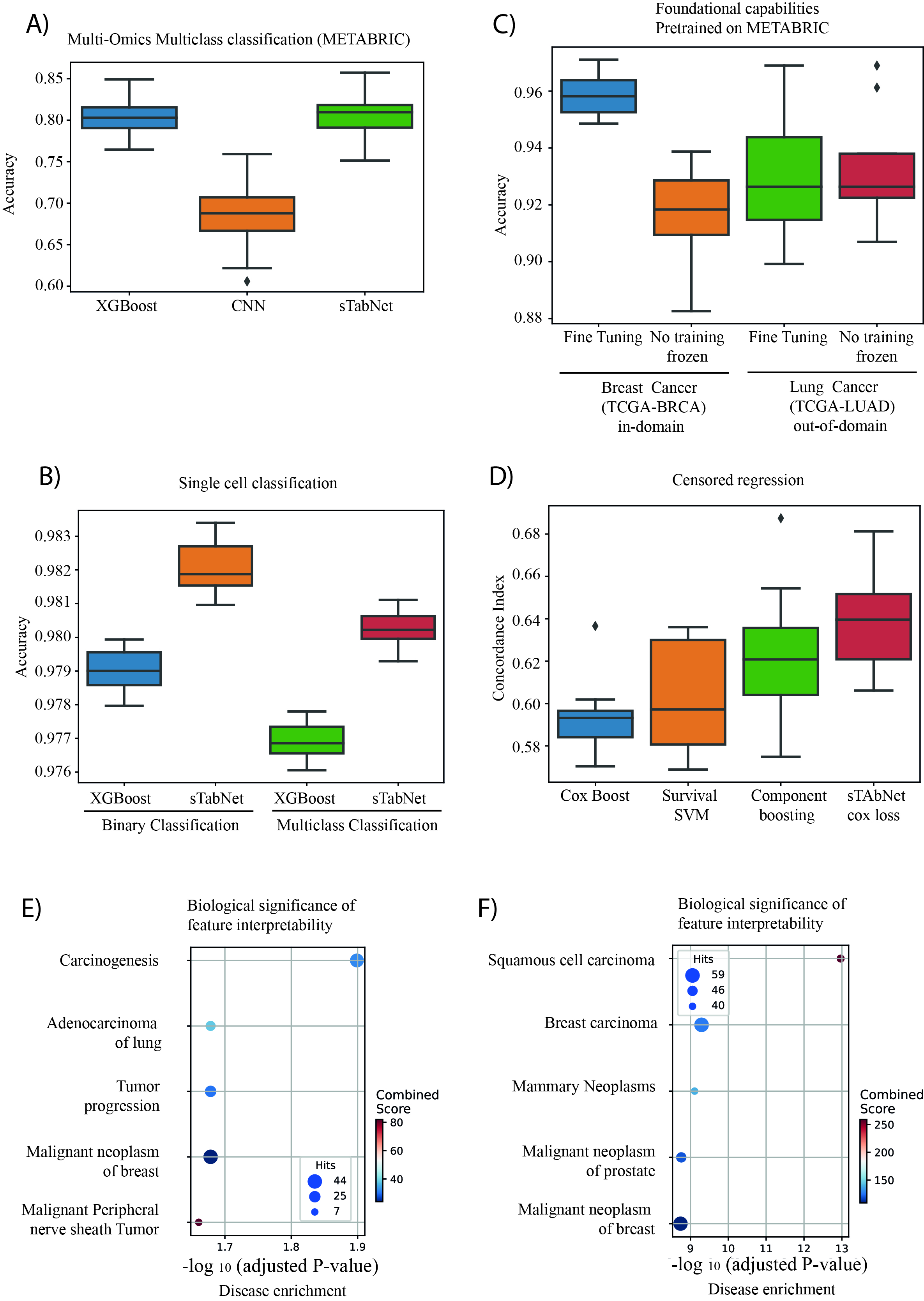}
        \caption{  sTabNet provides a foundational model to perform in-domain and out-of-domain fine-tuning, it is interpretable and outperforms tree-based models in general.\textbf{A} Comparative table between XGBoost and sTabNet and CNN on METABRIC multi-omics dataset (multiclass classification). \textbf{B} in-domain (breast cancer) and out-of-domain (lung cancer) adaptation. The model was trained on the Metabric dataset, then fine-tuned on other datasets. Accuracy on TCGA-BRCA (same domain of the METABRIC dataset) and  TCGA-LUAD (different domain from original dataset) for fine-tuning or feature extraction. \textbf{C} Binary and multiclass classification accuracy for sTabNet and XGBoost on single-cell data (breast cancer (GSE161529)). \textbf{D} Concordance index for METABRIC survival analysis.   
         \textbf{E}. Disease enrichment for the  100 top genes according to attention importance. \textbf{F}. Disease enrichment for the  100 top genes according to attention importance from METABRIC.
\textbf{change here }  \textbf{B}   Multiclass classification accuracy for XGBoost, CNN, and sTabNet  (METABRIC dataset).   }
        \label{fig:3}
    \end{figure}

\subsection{sTabNet vs. tree-based models for tabular data}

Although different transformer-based models have been tested, different researches suggest that the high capacity of neural networks hinders their applicability to tabular tasks \cite{kadra2021welltuned}. Since information about the features is often not available, we described a method to leverage sTabNet to address problems in arbitrary tabular datasets. In fact, the aim was to identify a simple method to impose sparsitybefore training while being competitive with tree-based models (Fig. 4A). Despite not conducting a hyperparameter search and using a simple architecture, the sTabNet is shown to outperform the tree-based model: median accuracy 0.71 versus 0.70 (Fig. 4B and supplementary table 5, and Fig. S7-8).

We also checked if our random walk approach could be used for complex datasets such as genomic datasets. We obtained similar results with Gene Ontology when using the random walk-based approach, showing that this approach is scalable and can be extended to any dataset (Fig. 3F, S9 and supplementary table 6).

Within a sparse neural network layer, each neuron's connections are determined by a random walk across a feature graph. This process ensures that the neuron establishes connections only within a localised, harmoniously related subset of features (i.e., its neighbourhood) on the graph. We noticed that the top random walk (selected using an additional attention layer) always represents the same neighbourhood (Figs. 4C - D). We conducted an ablation study on the features by removing the top 5 features to verify and study the effect of these features (in connection with their neurons). Fig. 4E shows that the removal of the involved feature impacts the accuracy. Interestingly, removing these features increases the number of false positives in comparison to random feature removal (Fig. 4F). We did not notice an increase in false negatives with feature ablation (Fig. 4G). These results suggest that local patterns in the feature graphs can be associated with a particular class and thus the emergence of modularity.
The recall ability of a model is dependent on the interconnectivity of its features collectively, while the precision is dependent on the existence or absence of a certain set of features in a pattern. Our results show consistently that the model is able to identify these features that are important for the ability to precisely classify a pattern which contributes to a better precision.

\begin{figure}[h!]
        \centering
         \includegraphics[width=13cm, height=13cm]{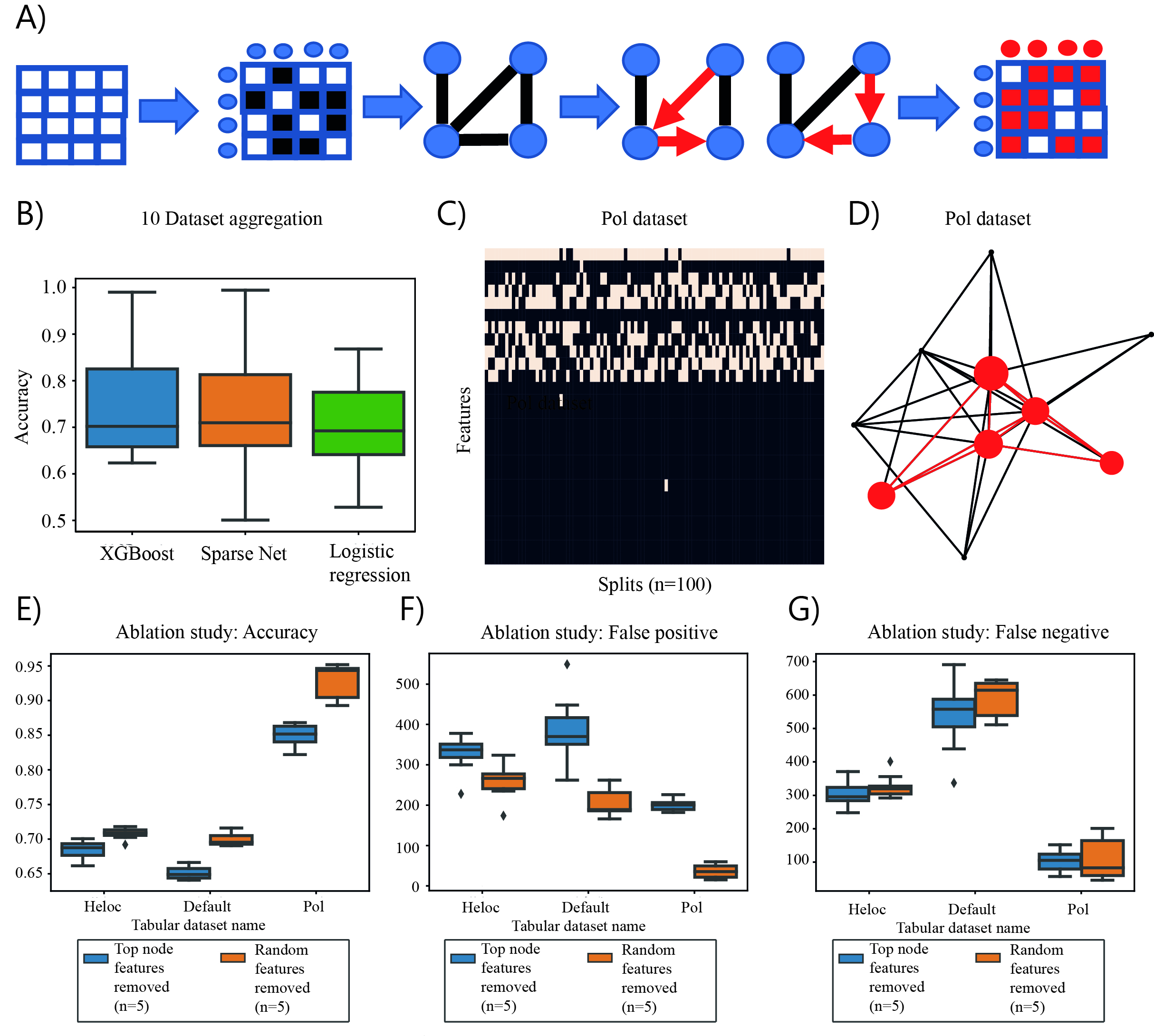}
        \caption{ Generality of the proposed sTabNet architecture. sTabNets are competitive to tree-based models for tabular data. \textbf{A} The algorithm that involves a random walk process is used to adapt the model to any tabular dataset. We calculated the similarity between the dataset's features (cosine similarity matrix) and used the obtained matrix to generate a feature graph. We conducted random walks on the feature graphs to explore the local neighbourhood of each feature. We use that knowledge to build the sparse matrix of the neural network  \textbf{B} Binary classification accuracy for the tabular benchmark (100 models for 3 techniques: we generated 10 models corresponding to 10 runs -each with a random training/testing split- for 10 datasets). \textbf{C} Feature presence in the top random walks (100 experiments on the pol dataset-each with a random training/testing split) \textbf{D} Feature graph for the pol dataset (isolated nodes are removed), the red highlightted nodes represent the top 5 features present in the top random walk. \textbf{E, F and G} Ablation study: accuracy (E plot) or false positives (F plot) or false negative (G plot) performance for three datasets removing 5 random features vs. removing the 5 top features of the random walk process (10 experiments for each dataset).}
        \label{fig:4}
    \end{figure}

\section{Discussion}
Tabular data are ubiquitous, and yet they have been left behind by the artificial intelligence revolution \cite{Borisov_2022, shwartzziv2021tabular}. However, many crucial applications rely on tabular data (medicine, psychology, finance,  user recommendation, cybersecurity, and customer churn prediction)\cite{Borisov_2022, ulmer2020trust, clements2020sequential, ahmed2017survey, buczak2015survey, urban2021deep}. We argued that we do not need large models for these small datasets and that unduly many parameters are deleterious for tabular data. For this reason, we proposed a simple but powerful model in which we exert prior knowledge (if present) to impose sparsity in the network architecture a priori.
In this work, we used attention mechanisms to identify features' importance in neural networks. Using a synthetic dataset that we defined its features' importance beforehand to provide ground truth, we showed how attention importance allows us to separate the informative features from noisy features swiftly. When this idea is applied to real-world datasets where we have domain knowledge (like cancer datasets), it will enable the identification of important features and can be used to explore insights about the dataset. This is important in many contexts (from biological to finance datasets and more), where knowing which feature is important is essential. We also showed how to \textbf{define} our sTabNet model when domain knowledge is lacking and we showed that the process is reliable and produces consistently good results for different benchmarks.

In addition, we showed that it is vital to consider data where the ground truth about feature importance is known when evaluating an algorithm's interpretability. We have highlighted how even established methods are not always reliable, especially for complex tasks with numerous features. Moreover, attention weights can be used for feature selection to eliminate non-informative features. These factors come in handy for real-world datasets, which are often noisy and have many redundant features.

Most previous comparisons between tree-based models and neural networks have been conducted on simple datasets with a limited number of features \cite{grinsztajn2022treebased}. These datasets are not representative of real-world scenarios as biologically complex datasets. We showed that when it is important that the model is learning a complex data representation, sTabNet are competitive with tree-based models. Moreover, in these complex tasks, neural networks can be used to extract or select features, perform transfer learning, and thus end-to-end training, establishing the foundational ability of our proposed sTabNet model. All these tasks are common to complex datasets in different fields (from medicine to customer churn prediction), and this opens exciting opportunities and applications. 

While other sparse models have been proposed when there is external knowledge about the features of the data, we defined an approach to define sTabNet for any tabular dataset. We showed that sTabNe has better results than tree-based models. Perhaps, as suggested before, for datasets with few features, the dense counterparts have too much capacity for the task. Constraining the neural network with sparsity in a systematic unsupervised learning manner is effective in allowing the model to learn the correct patterns.
Moreover, this method forces neurons to learn about the local neighbourhood of a feature in the feature graph, making the neural network more interpretable. Interpretability is essential for the applicability and adoption of neural network models.

However, our method also generates isolated nodes; in the future, other alternatives to developing the feature graph could be explored. For example, alternative distance measurements or a weighted Node2vec random walk can be explored. In addition, L1-regularization (or other regularization techniques) can be used to increase the network's sparsity.

\section{Material and methods}

\subsection{Tabular attention mechanism and sparsity layer definition}
We intend to build a neural network with sparsity and attention mechanisms. We define the attention mechanisms as an adaptation of the mechanisms described for text sequences \cite{bahdanau2016neural, luong2015effective, mnih2014recurrent, vaswani2017attention} with two principles: they should be adapted for tabular data and naturally define the features importance of such data. Moreover, to increase the expressive power of our established models, we equip them with a linear transformation learning of the input (simply multiplying with learnable weight vectors).

Given a layer $l$ with a 2D weight matrix $\boldsymbol{W} \in \mathbb{R}^{n \times m}$ representing the weights connecting $n$ neurons in layer $l$ to $m$ neurons in layer $l+1$. Let $\boldsymbol{X} \in \mathbb{R}^{b \times n}$ be a tabular dataset, where $n$ is the number of features and $b$ is the number of data points (or batch size). We consider a binary mask matrix $\boldsymbol{A} \in \{0,1\}^{n \times m}$ that determines which connections are kept or removed. A sparse weight matrix for the layer $l$ is then defined as the Hadamard product $\boldsymbol{A} \odot \boldsymbol{W}$. 
Therefore, the output of the \textit{first} layer of our sTabNet architecture is 

\begin{align*}
    \boldsymbol{H}^1 = \sigma(\boldsymbol{X}(\boldsymbol{A} \odot \boldsymbol{W}))
\end{align*}

where $\sigma$ is a non-linear activation function (e.g., ReLU, sigmoid), see Fig 1. B.

We introduce the attention mechanism as follows. Let $w$ be a learnable vector of size $n$, and the attention vector is given by 

\begin{align*}
   \alpha = \text{softmax} (\varphi (\boldsymbol{X} , w))
\end{align*}

where $\varphi$ is one of the functions given in Table 1 \cite{bahdanau2016neural, luong2015effective, mnih2014recurrent, vaswani2017attention}. 

\begin{table}
        \caption{ Score functions for attention mechanism}
            \begin{tabular}{r@{\quad}rl}
                \hline
                \multicolumn{1}{l}{Mechanism}&\multicolumn{2}{l}{score function ($\varphi$)}\\
                \hline
                Bahdanau inspired \cite{bahdanau2016neural}  & $\text{tanh}(\mathbf{X} \cdot \mathbf{w})$ & \\
                  dot-product inspired \cite{luong2015effective} & $\mathbf{X} \cdot \mathbf{w}$& \\
                content-based inspired \cite{mnih2014recurrent}  &   $\text{cos}(\mathbf{X} \cdot \mathbf{w})$ & \\
                scaled dot-product  inspired \cite{vaswani2017attention} &   $  \frac{\mathbf{X} \cdot \mathbf{w}}{\sqrt{n}}$& \\
                \hline
            \end{tabular}
    \end{table}

    As defined in \cite{kipf2017semisupervised}, every layer of a neural network can be written as a nonlinear function, where $\boldsymbol{A}$ is the adjacency matrix of a graph. In our model,  $\boldsymbol{A}$ is a matrix that controls the connections among neurons of the input (features) and hidden layers. It could also be interpreted as an adjacency matrix of a graph encompasses these neurons as nodes. The edges are interactions between the features and hidden neurons (Fig. 1C) . 
    Thus, the attention score is defined as a value of the node's importance in the feature graph.

    \subsection{Analysis of the interpretability of the sTabNetworks}

    To test if the attention mechanism score can capture the importance of a feature in the dataset, we built simulated data for a multi-classification task. The multiclass dataset is built as in \cite{Guyon2003DesignOE} and using scikit-learn package implementation,  where the difficulty of the classification task is regulated by a hyperparameter (separation coefficient, range 0.1-1). Where 0.1 is the most complex task for the model and is a trivial separation task. We defined informative and not informative features (random noise) in the dataset. We used scikit-learn standard implementation (\textit{make classification}), resulting in a matrix $\boldsymbol{X} \in \mathbb{R}^{b \times n}$  (where n is the sum of informative and non-informative features). The sTabNet and XGBoost have been then trained to optimise a multi-classification objective (6 classes), i.e. categorical cross-entropy as a loss function.

    For each separation coefficient, we trained ten different models (1000 examples with 100 features of which 10 informative and 90 non-informative) and measured the classification accuracy on the test set (0.2 of the dataset). We then examined whether a model can discriminate and separate informative features from random noise. For this purpose, we analysed the importance score (for XGBoost) or the attention weight assigned to each feature in the sTabNet. For XGBoost, we also used the SHAP value to determine features importance \cite{NIPS2017_8a20a862}. For the sTabNet, we analysed whether the attention weight can be considered a feature importance measure. As defined in \cite{tomsett2019sanity}, we conducted Least Relevant First (LeRF) and Most Relevant First (MoRF) studies to assess the fidelity of the feature importance attribution.

    \textbf{Technical details.} For XGBoost and SHAP, we used the standard parameters as defined in the XGBoost package. For the sTabNet, we used a simple architecture in Keras. The architecture has an input layer, attention mechanism 100, a sparse layer with 100 neurons, a linear layer with arbitrary 64 neurons (to test the effect of attention only not the effect of domain knowledge or unsupervised learning), and an output layer with softmax activation). We used dropout regularisation, ADAM optimiser, and categorical cross-entropy as a loss. Since previous works focused on extensive hyperparameter optimisation \cite{kadra2021welltuned} for the tabular neural networks, we intentionally did not conduct it to show performance with simple settings. For each separation coefficient, we randomly split 100 times (100 times cross-validation) and calculated the multiclass accuracy (both for XGBoost and the sTabNet). We extracted the feature importance attributed to each feature from the model.

    For LeRF and MoRF analysis, we iteratively removed the most or least important feature according to the mean importance (mean of 10 models) and retrained on the reduced feature set.
    We also checked the stability for XGBoost and sTabNet by progressively increasing the number of noisy features (while keeping the number of important features constant to 10) and analysing the importance attributed to the features (informative and non-informative).

    \subsection{Application to multi-omics dataset, fine-tuning, and feature extraction}
    Multi-omics data are notoriously complex to handle: they are expensive to collect, hetero-modals, and suffer the curse of dimensionality (for a dataset $\boldsymbol{X} \in \mathbb R^{m\times n}$, we have $n >> m$) \cite{reel2021using}. While designing an algorithm for this data is challenging, identifying important features leads to direct important utilisation \cite{reel2021using}.
    
    We used the following datasets: METABRIC \cite{curtis2012genomic} (RNAseq and mutations), TCGA-BRCA, and TCGA-LUAD \cite{cancer2013cancer} (only RNAseq data for both). We initially trained our sYabNet model with the above-defined attention mechanisms on METABRIC, testing different activation functions and comparing it with XGBoost, a fully connected neural network model (FCNN), and a Convolutional Neural Network model (CNN). We then tested its in-domain and out-of-domain capabilities with and without fine-tuning.

    \textbf{technical details} For the METABRIC dataset: We used XGBoost with a multiclass objective as suggested in the official library. The sTabNet was built as before but using Gene Ontology \cite{Ashburner2000} to define the sparse matrix $\boldsymbol{A}$ for the sparse layer. Given the more complex nature of the dataset, we added an extra layer at the end to increase the complexity of this sTabNet model. 
  
We used dropout as regularization. The sTabNet model was trained for 20 epochs and batch size 1024. We then used a 100-fold split for evaluation. Disease enrichment was conducted on the top 100 genes according to attention importance (ranked average of the 100 experiments). We used the GSEApy Python package and DisGeNet's disease database.  
   To show the fundaiotnal capabilties of our sTabNet architecture, a sTabNet model was pre-trained on the METABRIC dataset to be used with and without fine-tuning. For feature extraction, we removed the last two layers from the frozen models and extracted the features for the TCGA-BRCA and TCGA-LUAD datasets. On the extracted features, we built a simple logistic regression model for a binary classification task.
    For the fine-tuning protocol, we added two fully connected layers to the frozen model. The model was fine-tuned using binary cross-entropy as a loss function. The plots show the results for 10-fold cross-validation.

    \subsection{Sparse net application to single-cell RNA seq dataset}
    We used single-cell data from the Tumor Immune Single-cell Hub 2 (TISCH2) of breast cancer (GSE161529). We used annotated data with minimal preprocessing as suggested by ScanPy \cite{wolf2018scanpy}. We used sTabNet architecture as above to train two fresh networks for binary classification (tumour/normal cell) and multi-classification tasks (cellular type prediction). sTabNet was built for the METABRIC dataset, and we used XGBoost, as described above. The sTabNet model was trained for 20 epochs and batch size of 1024. We then used a 10-fold random split for evaluation (a 10-fold split for evaluation was also used for XGBoost). We used a typical binary and multi-class final layer, i.e. for binary classification, we used a sigmoid final layer and binary cross entropy as a loss, while for multiclass classification, we used a softmax layer and categorical cross-entropy.

    \subsection{Sparse net application to survival analysis} 
    For survival analysis, we used the METABRIC dataset as described above, we used the associated metadata for overall survival. We modified the sTabNet to adapt to the survival prediction task.  Specifically, we used the Breslow approximation as a loss function \cite{yang2022fastcph} for the last layer and measured the performance with a concordance index. We compare with Scikit-survival library \cite{sciki_surv}, which implements different algorithms for survival analysis. For this purpose, we selected two different gradient-boosted models (Gradient Boost and Component Wise Gradient Boost) and the fast extension of the support vector machine for survival analysis. For each algorithm, we conducted a 10-fold random split validation. 

\subsection{ Generalisation of sTabNet for tabular datasets with no-domain knowledge  }

In biological terms, a pathway is an approximation of locally connected features that interact among themselves to fulfill a biological function \cite{wysocka2023systematic}. These could be seen as a group of connected nodes in the feature graph (ex. a group of proteins in the protein-protein interaction graph). In this section, we will deal with the more general case when we do not have external knowledge of feature interactions, which is the case in most tabular datasets domains. We wanted to extend sTabNet to other domains where we need to enforce sparsity but there is no knowledge available about the features or their interactions. We, therefore, hypothesized that this structure could be approximated as a random walk starting from each feature and walking in the feature graph, using Node2vec biased random walk \cite{grover2016node2vec}. In this way, we can explore the locality of each feature and their interactions. 

Let $\boldsymbol{X} \in \mathbb R^{m\times n}$ an input dataset we define the feature graph ${G}$ isand its adjacency matrix $\boldsymbol{M} \in \mathbb R^{n\times n}$ in the following way. The nodes in $G$ represent $\boldsymbol{X}$'s features, and the edges exist if there is a similarity between two features $i \in \mathbb n$ and $j \in \mathbb n$. In other words, if the cosine similarity between the features $i$ and $j$ in $\boldsymbol{X}$ is higher than 0.5 or less than -0.5, $M_{ij} = 1$, otherwise $M_{ij} = 0$. We then perform $r$ Node2vec random walks of $t$ steps for each node in the graph $G$ (we did a parametric search for $r$ and $t$ and found that three and five are the best values respectively). We considered a matrix $\boldsymbol{A} \in \mathbb \{0,1\}^{n\times r}$, where $n$ is the dataset features and $r$ is the number of the random walks. For the random walk $j$ if the node $i$ is present in the random walk $j$, $A_{ij} = 1$, otherwise $A_{ij} = 0$. This process is described in Fig. 1 C-D and Algorithm 1.
    
    \vspace{10pt}
\begin{algorithm}[H]
\SetAlgoLined
\SetKwInOut{Input}{Input}
\SetKwInOut{Output}{Output}
\Input{Dataset $X$ with dimensions $R^{m\times n}$}
\ output {matrix of random walks}
Calculate cosine distance matrix $D$ for features $n$\;
Initialize an empty graph $G$\;
\For{$i$ in range($n$)}{
  \For{$j$ in range($n$)}{
    \If{$D[i,j] > 0.5$ or $D[i,j] < -0.5$}{
      Add an edge between nodes $i$ and $j$ in $G$, if $i \neq j$ \;
    }
  }
}
\For{$i$ in range($n$)}{
  \For{$r$ in range($3$)}{
    $v \gets G[i]$\;
    \For{$t$ in range($5$)}{
      Node2Vec RandomWalk()\;
    }
  }
}

\caption{Generate graph and perform random walks}
\end{algorithm}

In a classical fully connected feed-forward neural network, each neuron can be considered a walk that connects this neuron to each node in the feature graph (Fig. 1C), thus learning a \textit{global} approximation for the whole feature graph. Instead, a node in the sparse layer of our sTabNet, obtained by a random walk, is connected to relevant nodes and thus is learning \textit{local} approximation for the feature graph. Moreover, since we are using Node2vec we can tune and control how big its locality is and how much of the locality of a feature we want to explore. Each neuron of the network is specializing in a certain set of features (locality of the feature graph). This process reduces the impact of noise and irrelevant features. In fact, this enforced sparsity is making the model more robust to learning spurious correlations, which is a main cause of overfitting.
    
\textbf{Training details.} To test our hypothesis that random walk is a good approximation for domain knowledge connectivity, we used the benchmark datasets described by \cite{grinsztajn2022treebased} and focused on datasets with more than 20 features and less than 100 K examples. We built the sparse matrix $A$ as described above; we performed three random walks of size 5 for each feature. We compared the sTabNet accuracy with XGBoost and logistic regression. Figure 4 shows the result of the training.

We also performed an ablation study. We added another attention layer \textit{after} the sparse layer to study which neuron is more effective for our architecture. We extracted the attention weight (associated with each neuron) and conducted 100 different splits to account for the random walk variability, tracking which nodes were associated with the highest attention weight. We then removed these features,measured the performance and compared it with the model performance when we removed five features at random. The results are shown in Figure 4 E-G, where we can see that the performance deteriorated more when random walk features were removed demonstrating that these random walk-identified features are more important than other features. In other words, our study shows that the random walk process is effective in identifying the most important connections between nodes and features making our architecture well-suited to the problem at hand without using domain knowledge.

 \subsection{Data availability}

No unique data were used in this study. All the data used are either published or generated by simulation. The process is based on a standard library (scikit-learn) for simulated data and is described in detail in the methods. For the other experiments, we used only published data, and data are publicly accessible from the provided references. 

    \subsection{Code availability}

We are providing the code for the sparse neural network. As new implementations of the sTabNet become available, we will include them in the repository.  Code can be accessed at: https://github.com/SalvatoreRa/sTabNet.

\clearpage



\nolinenumbers
\section{Author contributions}

SR performed the experiments, acquired data, and analyzed the results. NJ provided datasets for the analysis. SR and AA wrote the manuscript. SV and SG revised the manuscript and acquired funds. SR and AA designed, conceptualized and supervised the study. All authors contributed to the article and approved the submitted version.

\section*{Acknowledgements}

We want to thank Jonathan Schmiedt and Thomas Wursten, members of the OPM IT team, for their technical assistance during this project.

\section{Conflict of Interest}

The authors declare no competing interests.


\section{Supplementary methods}

\subsection{ Different sTabNet architecture details}

We are providing here additional information about the technical details of the new tabular foundational model used in this work. For the experiments (sections 4 and 6), the sTabNet was defined as shown in the following table:

\begin{table}[ht]
\centering
\begin{tabular}{ |c|c|c| }
\hline
 Layer & Units & Other information \\ 
 \hline
 Attention Layer & - & scaled dot type \\
 (optional) Sparse layer & 100 & tanH activation - Omics fusion  \\
 Sparse layer & 100 & tanH activation   \\
 Dense layer & 64 & ReLU activation and dropout regularization (0.3)   \\
 Dense layer & n & according to the task (ex. softmax, sigmoid or linear)\\ 
 \hline
\end{tabular}
\caption{Scheme of sTabNet}
\end{table}

For the syntethic data (section 4) the sTabNet was defined without the optional layer and with softmax where $n=6$.  For the multi-omics experiments (setion 5), since it is multimodal (DNA and RNA at the same time) the sTabNet was defined with the optional layer with a final softmax activated layer where $n=6$. For the censored regression, we had also the optional layer,  with a final linear activated layer, where $n=1$. In section 6, we used the model without the optional layer and with final sigmoid activated layer, where $n=2$ (binary classification).


\subsection{Attention Mechanisms as a measure of feature importance}

Here we are showing the results using the other three attention mechanisms on the same synthetic dataset (section 4). The architecture and dataset are the one shown in figure 1, only the calculation of the attention is different. All the attention mechanisms are behaving in a similar way (figure S1).

Additional training increases the performance of the neural network when the classification task is complex (figure S2). Additional epochs are increasing the accuracy of the models and a better separation between the weights associated with informative features and non-informative features (showing that increasing the number of epochs is helping the model to learn a better representation of the data). We increased the number of epochs from 200 to 500.

\subsection{Consensus with other commonly used algorithms}

We calculated the feature importance with well-known methods (XGBoost feature importance, Scikit-learn permutation importance, LIME, Shap Value, Logistic regression coefficients) and we compared with sparse net (we used the attention weights as a measure of feature importance). For each model, we did 10-cross-fold validations, average obtained values, and reported in the table the ranked importance of the features for the adult Census Income dataset (a well-described dataset where the meaning of the features is known).

We observed a commonality among the different methods (sparse net included) for the most important features and the least important features (Table 3).

\begin{table}[h!]
        \caption{ Ranked featured importance (10 splits) for Adult Census Income dataset }
            \begin{tabular}{|c|c|c|c|c|c|c|}
                \hline
                Feature & XGBoost &	Permutation &	LIME &	Shap &	Logistic  &	sparse net \\
                  &  & importance &	 &	 &	regression & \\
                \hline
                 age & $$6$$ & $$4$$ & $$1$$ & $$3$$ & $$4$$ & $$3$$ \\
                 \hline
                 capital gain & $$2$$ & $$1$$ & $$4$$ & $$1$$ & $$1$$ & $$1$$  \\
                 \hline
                 capital loss & $$5$$ & $$6$$ & $$10$$ & $$11$$ & $$5$$ & $$4$$ \\
                 \hline
                 education & $$14$$ & $$13$$ & $$13$$ & $$14$$ & $$12$$ & $$13$$  \\
                 \hline
                 education num & $$3$$ & $$2$$ & $$5$$ & $$2$$ & $$2$$ & $$7$$  \\
                 \hline
                 fnlwgt & $$12$$ & $$11$$ & $$8$$ & $$12$$ & $$9$$ & $$8$$  \\
                 \hline
                 hours per week & $$9$$ & $$8$$ & $$7$$ & $$5$$ & $$3$$ & $$6$$  \\
                 \hline
                 marital status & $$4$$ & $$7$$ & $$3$$ & $$7$$ & $$6$$ & $$2$$  \\
                 \hline
                 native country& $$13$$ & $$12$$ & $$14$$ & $$10$$ & $$13$$ & $$12$$  \\
                 \hline
                 occupation & $$7$$ & $$5$$ & $$6$$ & $$8$$ & $$14$$ & $$10$$  \\
                 \hline
                 race & $$11$$ & $$14$$ & $$12$$ & $$9$$ & $$11$$ & $$14$$ \\
                 \hline
                 relationship & $$1$$ & $$3$$ & $$2$$ & $$4$$ & $$10$$ & $$5$$ \\
                 \hline
                 sex & $$8$$ & $$10$$ & $$9$$ & $$6$$ & $$7$$ & $$11$$ \\
                 \hline
                 work class & $$10$$ & $$9$$ & $$11$$ & $$13$$ & $$8$$ & $$9$$  \\
                 \hline

                \hline
            \end{tabular}
    \end{table}

\subsection{Feature extraction for single-cell data}

We used single-cell data from the Tumor Immune Single-cell Hub 2 (TISCH2) of breast cancer (GSE161529). We used a sparse net for binary classification (tumor/normal cell) or multi-classification tasks (cellular type prediction).  For each model, we did 10 cross-fold validations, averaged obtained values, and reported the accuracy. After freezing the model we also extracted the data representation and performed data dimensional reduction with UMAP to plot it.
The dataset and associated metadata were downloaded from the TISCH2 database. The data were filtered and normalized using the Scanpy library. We defined the sparse net as described in the methods. For binary classification, we used a sigmoid final layer and binary cross entropy as loss while for multi-class classification we used a softmax layer and categorical cross-entropy. The model was trained for 20 epochs with a 1024 batch size. 
We extracted from the frozen model the dataset representation obtaining a 64-dimensional vector for each cell. On the vectorial representation, we conducted UMAP projection for obtaining a 2-dimensional visualization

\subsection{sparse net as feature selection method}
Genomic datasets suffer from the curse of dimensionality, where there are many more features than examples (many of these features are highly correlated). In many contexts, it is useful to have an algorithm to identify relevant features and apply it to filter out the redundant features. For this reason, we are showing a simple pipeline where we train a model to identify the relevant features using the attention weights of the sparse model. The identified features are then used to retrain another model (XGBoost) to show that the performance is not degraded when the redundant features are removed.
We trained the sparse net on the METABRIC dataset with the same architecture as described previously. We just add l1-regularization as regularization to each layer. We tried increasing values of regulation strength (1*10e-7, 5*10e-7, 1*10e-6, 5*10e-6, 1*10e-5, 5*10e-5, 1*10e-4, 5*10e-4, 1*10e-3, 5*10e-3). For each case, we trained 10-fold cross-validation on the training set and extracted all the non-zero coefficients for the attention weights. The features with non-zero attention weights were used to filter the dataset and train an XGBoost model for multiclass classification (10-fold cross-validation)

\subsection{Unsupervised approach for biological omics data}
In the previous experiments, we have shown how the attention mechanisms provide a feature-importance method in the context of neural networks applied to biological datasets. We have shown how the adjacency matrix is obtained using Gene Ontology's pathway membership, however, this approach is not exempt from limitations:
\begin{itemize}
      \item Pathway databases are generally manually annotated and therefore it is a time-consuming and expensive approach.
      \item They are generally incomplete. Relationships are derived from literature and each new update of the database can be published after a long time.
      \item They contain errors. Human annotation is error-prone and often errors are not corrected in a new version.
      \item Not all the genes are present in the databases or a member of some pathways. 
      \item Criteria for inserting a gene in a pathway can be subjective
    \end{itemize}

For these reasons, we decided to test if we could use an unsupervised approach to generate the pathway matrix. Here we present the comparison with two different approaches for the adjacency matrix generation (random walk approach on the features and protein-protein interaction based). We are also adding Reactome as a prior knowledge database for comparison. For these experiments, we used METABRIC (RNAseq and exome as described before) and as the baseline, the adjacency matrix was obtained from Gene Ontology (GO) as described before. We used a random walk approach on the constructed graph as described in algorithm 1. For protein-protein interaction (PPI) we built the PPI graph and for each gene, we conducted a random walk, we used the obtained random walk matrix for the sparse layer.

In the figure, we show that the random-walk-based approach and PPI-based approach lead to comparable results with the use of GO (average of 10-fold cross-validation). However, the Reactome-based neural network had a sensibly worse performance. For this reason, we excluded Reactome from the successive analyses. Moreover, attention weights for the three approaches (GO, random walk, PPI) are correlating and concordant. This shows that the random walk approach (or another unsupervised approach) is also suitable for biological datasets. Moreover, the PPI approach shows how we can adapt another biological dataset for analyzing biological data with a neural network. 

\textbf{Technical details}. GO experiment was conducted as described before. For the random walk implementation, we conducted a random walk for each gene (20 steps) and the obtained matrix was used in the sparse layer. For the PPI approach, the PPI graph was downloaded from the STRING database (https://string-db.org/) and we filtered out all the PPI with a combined score of less than 400 and all the genes not present in METABRIC. Then, we built the graph with the NetworkX library in Python, and we conducted a random walk for each gene (20 steps). For the Reactome experiment, we obtained the pathway's data from Msigdb (https://www.gsea-msigdb.org/gsea/msigdb) and we used the pathway as a matrix in a similar manner for GO. For all the experiments we used the same neural network architecture defined for Metabric experiments and we conducted 10-fold cross-validation. We extracted the for each experiment the attention weights. We used the Pearson correlation and Concordance index on the attention weights. We then enriched for the diseases (DisGenet) for the 100 top genes in the random-walk approach (rank according to attention coefficient).

\subsection{Comparison with clustering}

We have also tested the possibility of using clustering of the features instead of random walks. We decided to use the random walks implementation for the following motivations:
\begin{itemize}
      \item Very few clustering methods are generating overlapping clustering. This causes only features in the same clustering to interact with each other, leading to the potential loss of important interactions among the features (which could create an information bottleneck). While the core idea of sparse networks is to reduce the number, excessive pruning is known to harm performance. Moreover, other models exploit prior knowledge to introduce sparsity and we are here lacking this guidance. Random walks produce overlapping clustering instead. 
      \item Considering that we have no knowledge about the features random walks also introduce stochasticity. At the level of inductive bias, this creates a weaker inductive bias by less constraining the hypothesis space and thus allowing for better exploration of the hypothesis space. In fact, we have chosen node2vec because it allows the user to decide how much of the feature space is explored (deciding how much you want to move between breadth-first search and deep-first search approach).  
      \item Most of the success of decision-tree-based models derives from using them as an ensemble. In general, ensembles are as a much successful as they are diverse. Random walk naturally creates an ensemble, thus it is straightforward for a user to generate multiple models with random walks and combine the predictions of these models.  
      \item Since we are clustering the features of a dataset, the computational cost grows in function of the number of examples of the dataset. Moreover, clustering in multi-dimensional space becomes less efficient and accurate with the increase of dimensions. Instead, the random walk is conducted on the feature graph and it is independent of the number of the example.
      \item As we showed before, the use of the random walks approach has intriguing properties. In fact, it is forcing the nodes to learn a local zone of the feature graphs which can be connected to particular patterns.
      \item Lastly, as has been shown before a feed-forward neural network can be considered as a graph neural network (GNN) where the graph is fully connected. Thus, as shown before we can consider the attention weight (which we define above) as the importance of a feature in the feature graph. Therefore, we decided to use a random walk on the feature graph.
    \end{itemize}

\textbf{Technical details}. We selected the best hyperparameter combination ( number of random walks for a feature and the number of steps for random walks) for the random walk approach using a random search. We used k-means as a clustering approach, the number of clusters was selected using the elbow methods. We measured the accuracy of 100 splits to compare on the pol dataset.

\clearpage

\section{Supplementary Tables}
\vspace{\baselineskip}

\begin{itemize}
        \item \textbf{Supplementary table 1.} Ranked featured importance (10 splits) for Adult Census Income dataset
\vspace{\baselineskip}
      \item \textbf{Supplementary table 2.} Attention weights for METABRIC
\vspace{\baselineskip}

      \item \textbf{Supplementary table 3.} Enrichment terms for Metabric attention weights 
\vspace{\baselineskip}

      \item \textbf{Supplementary table 4.} Attention weights for single-cell data
\vspace{\baselineskip}

      \item \textbf{Supplementary table 5.} Accuracy and standard deviation (10 splits) on the tabular benchmark dataset for each algorithm.
\vspace{\baselineskip}

      \item \textbf{Supplementary table 6.} Enrichment terms for Metabric attention weights. We present the weights for Gene Ontology approach (GO),  random walk-based approach (RW), and protein-protein interaction approach (PPI)
\end{itemize}

\clearpage

\section{Figure Legend}

\begin{figure}[hbt!]
        \centering
         \includegraphics[width=14cm]{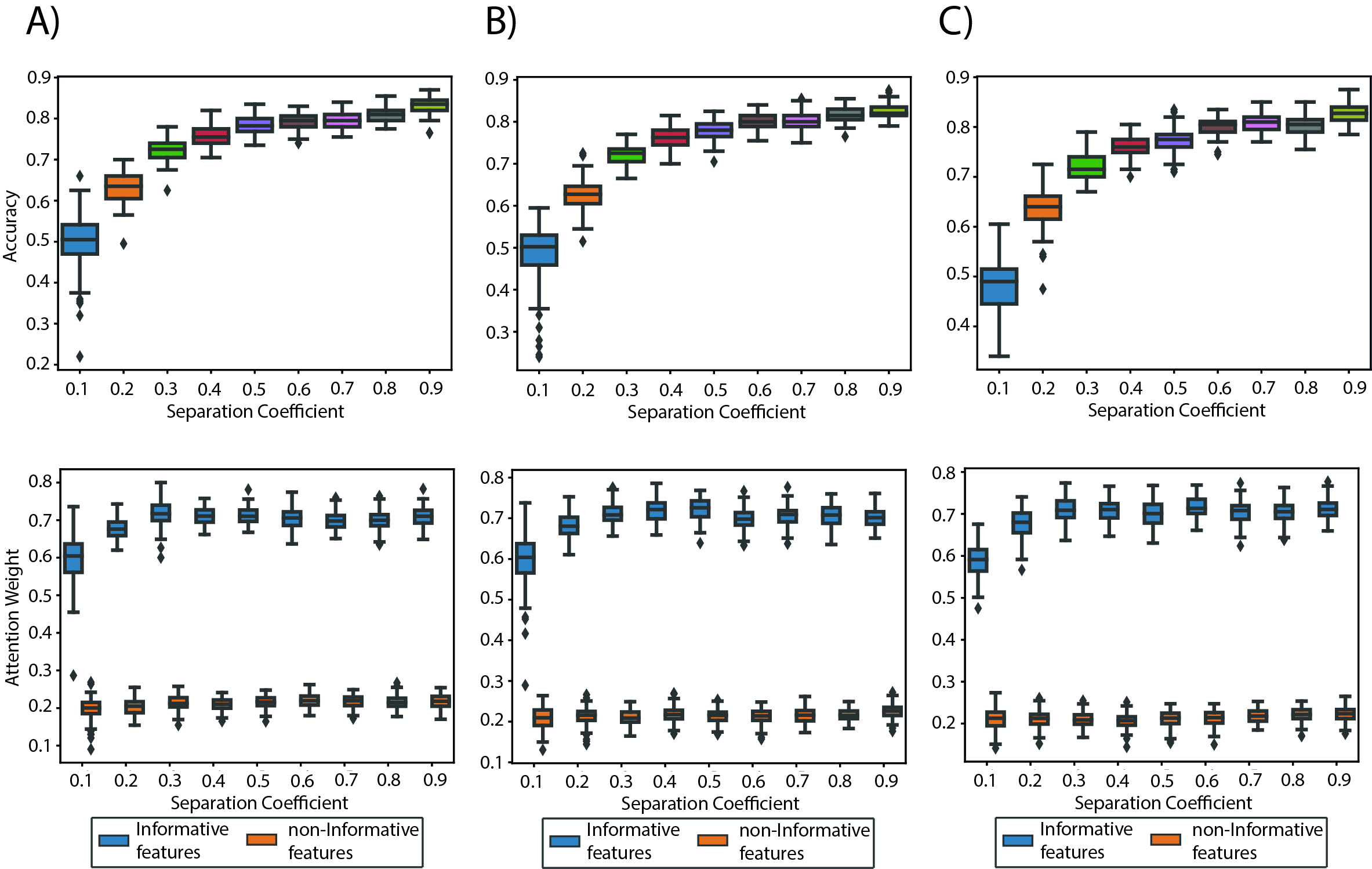}
        \caption{\textbf{Fig. S1} Attention mechanisms are a measure of feature importance. \textbf{A-C}. Each boxplot represents 100-fold hold-out validation, a lower coefficient represents a harder multiclassification task. The upper plot represents multi-classification accuracy in sparse net with an increase in separation difficulty. The lower plot represents the separation between the average importance weight (feature attention weight) assigned to real informative features and to not informative features.\textbf{A}. Bahdanau inspired attention mechanism. \textbf{B}. Luong-inspired attention mechanism. \textbf{C}. Graves inspired attention mechanism.}
        \label{fig:S1}
    \end{figure}

\vspace{\baselineskip}

\clearpage

\begin{figure}[hbt!]
        \centering
         \includegraphics[width=13cm]{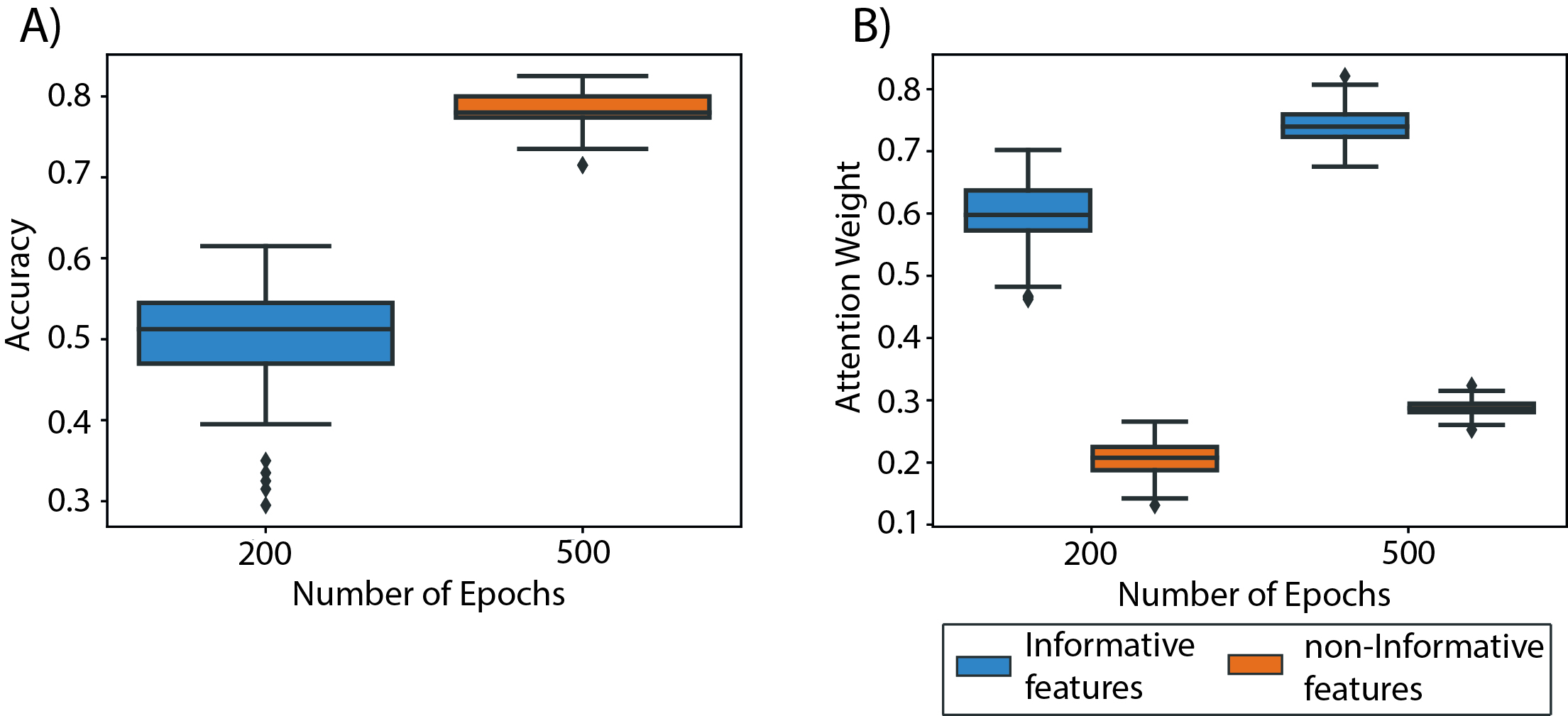}
        \caption{\textbf{Fig. S2} Attention mechanisms are a measure of feature importance. Sparse net trained for 200 or  500 epochs (0.1 separation coefficient), Accuracy (\textbf{A}), and separation attention weights for informative and non-informative features (\textbf{B}).}
        \label{fig:S2}
    \end{figure}

\vspace{\baselineskip}

\clearpage

\begin{figure}[hbt!]
        \centering
         \includegraphics[width=14cm]{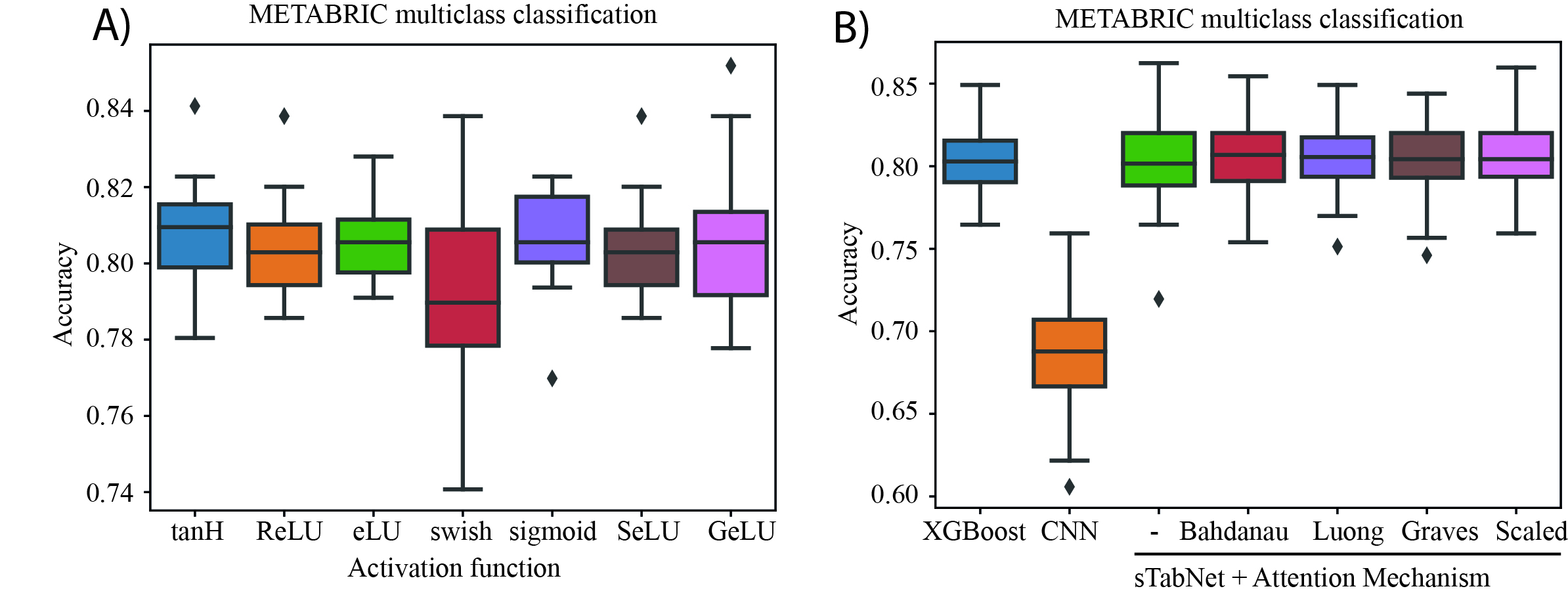}
        \caption{\textbf{Fig. S3} \textbf{A}. Multiclass classification accuracy for different activation functions (METABRIC dataset). \textbf{B}. Multiclass classification accuracy for different attention mechanisms  (METABRIC dataset). 
}
        \label{fig:S3}
    \end{figure}

\begin{figure}[hbt!]
        \centering
         \includegraphics[width=13cm]{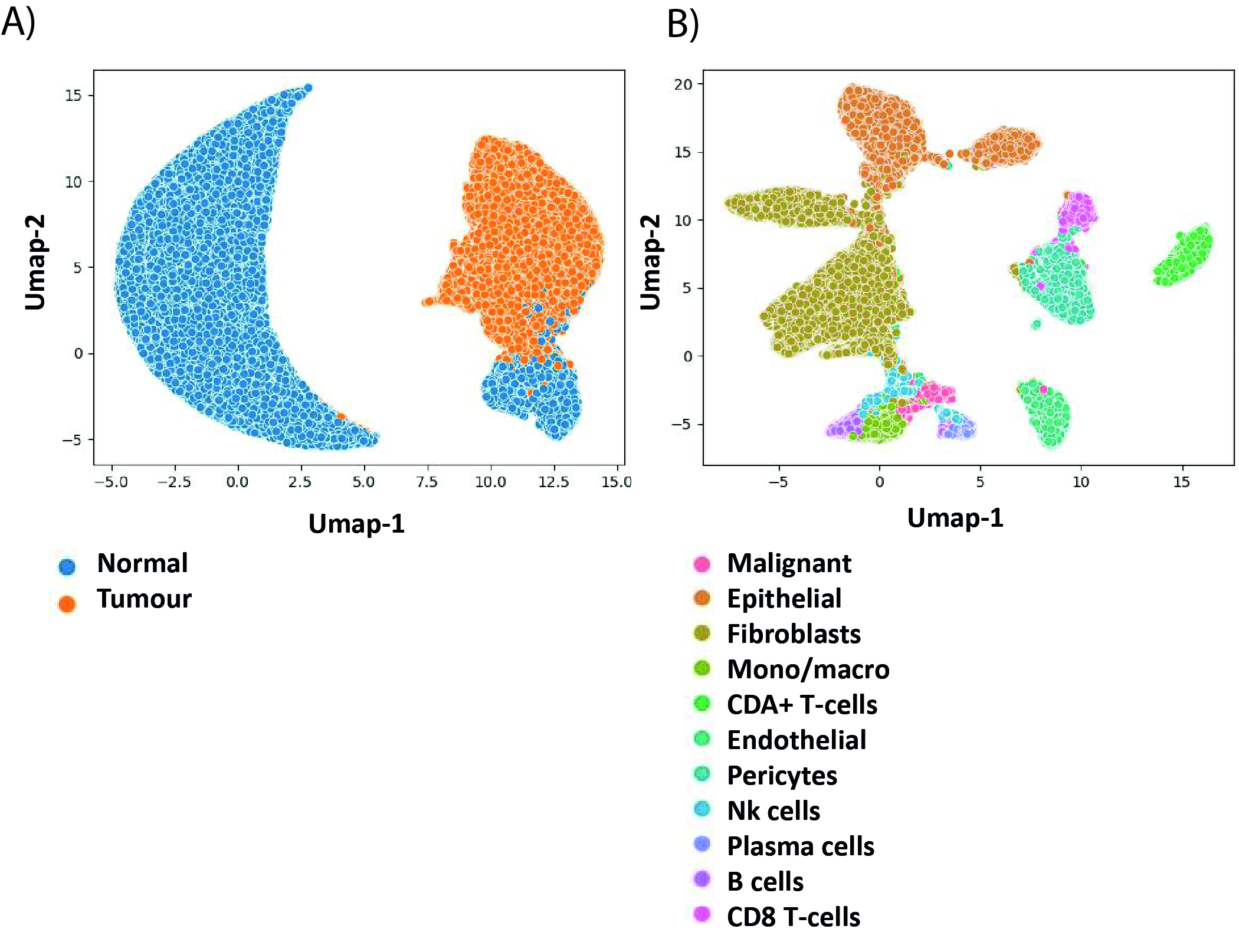}
        \caption{\textbf{Fig. S4} \textbf{A-B}) extracted representation for the sparse net with single-cell data. Attention mechanisms are a measure of feature importance. (\textbf{A}) UMAP projection for the binary classification task (\textbf{B}) UMAP projection for the multi-class classification task.
}
        \label{fig:S4}
    \end{figure}

\vspace{\baselineskip}

\clearpage

\begin{figure}[hbt!]
        \centering
         \includegraphics[width=13cm]{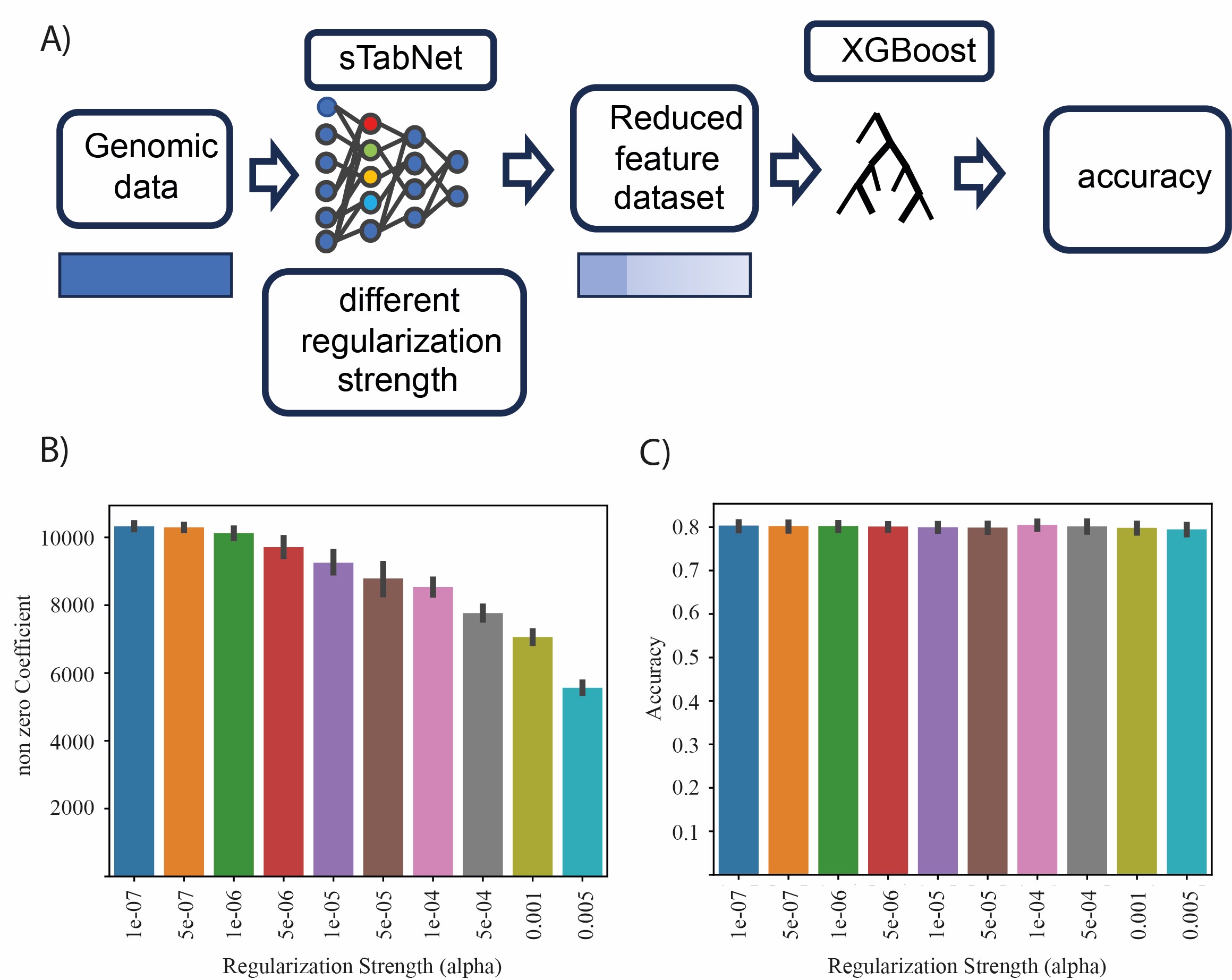}
        \caption{\textbf{Fig. S5}  Sparse net as feature selector. \textbf{A}  Briefly, we used the Metabric data and trained sparse net with increasing regularization strength (1*10e-7-0.005), using L1 regularization. We selected all the non-zero attention coefficients. We used the reduced dataset for training an XGBoost model.  \textbf{B} Number of features selected after training for different values of regularization strength (non-zero attention coefficients).  \textbf{C} Accuracy of XGBoost trained on the selected features in the previous step.
}
        \label{fig:S5}
    \end{figure}

\vspace{\baselineskip}

\clearpage

\begin{figure}[hbt!]
        \centering
         \includegraphics[width=10cm]{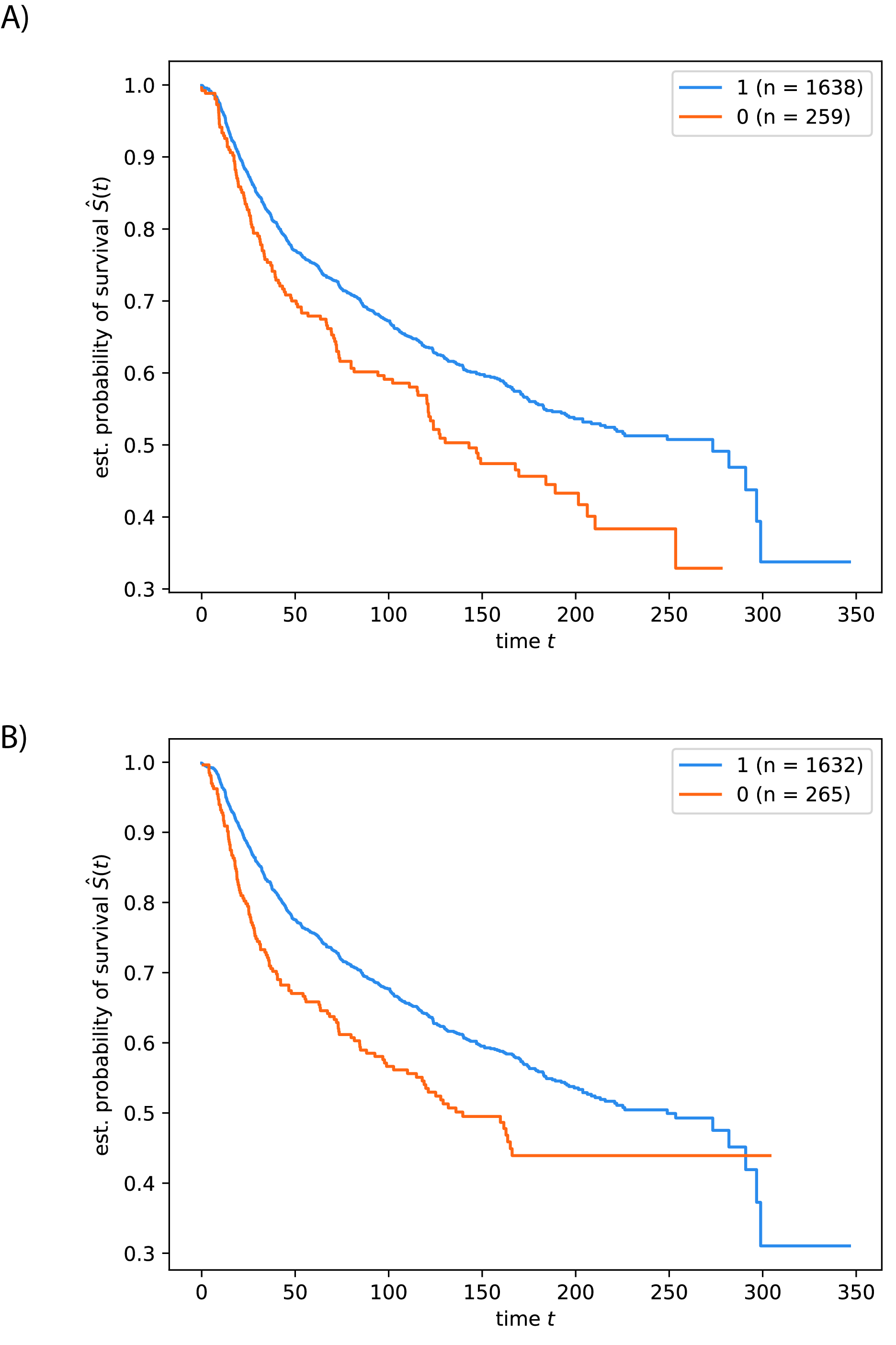}
        \caption{\textbf{Fig. S6} Survival curves for two selected genes as examples of the most important for survival. Top: RETN, down: MRPL2}
        \label{fig:S6}
    \end{figure}

\vspace{\baselineskip}

\clearpage

\begin{figure}[hbt!]
        \centering
         \includegraphics[width=13cm, height=12cm]{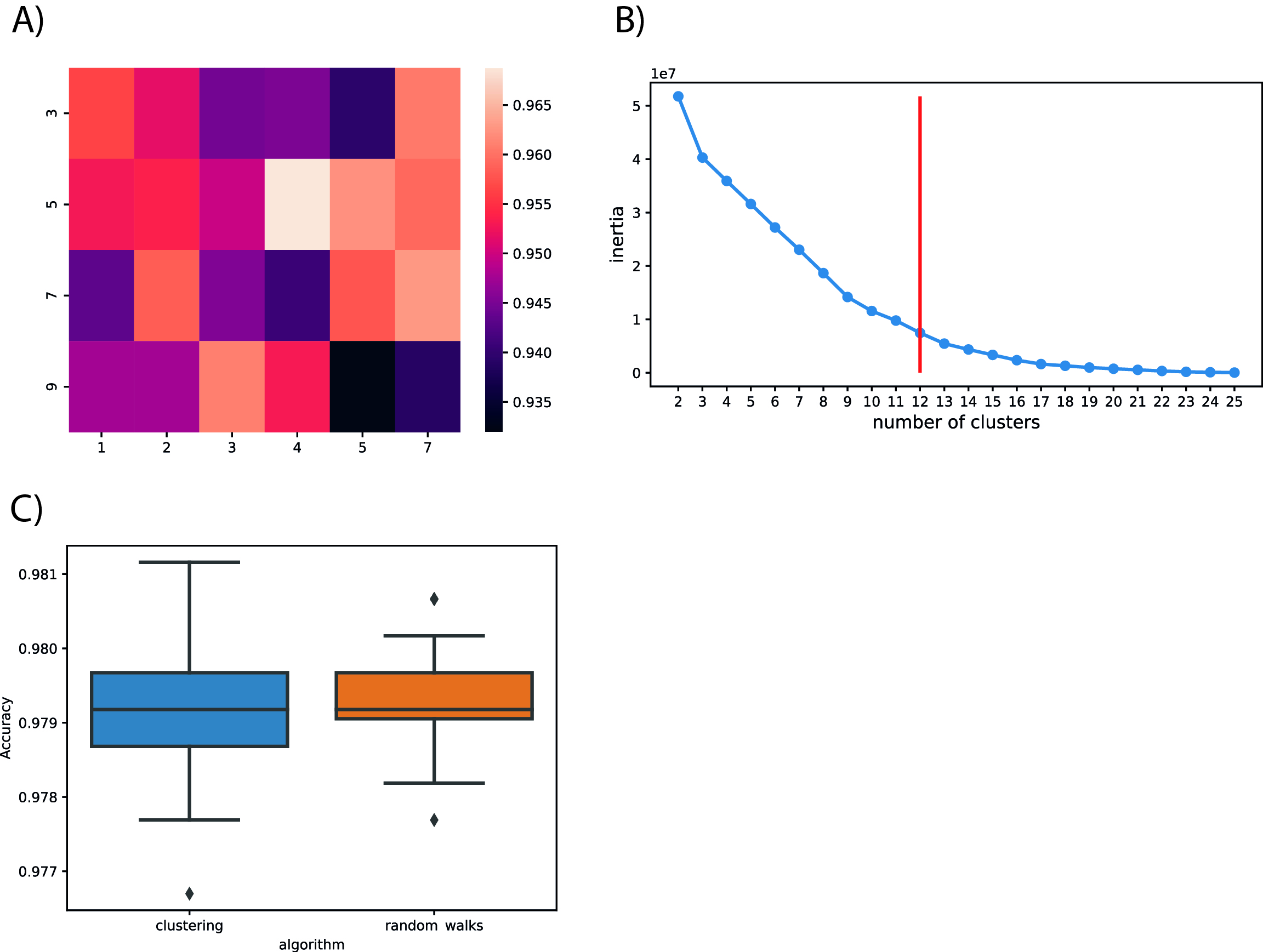}
        \caption{\textbf{Fig. S7} Comparison of clustering and random walk implementation. \textbf{A}.Random search for the best hyperparameter combination ( number of random walks for a feature and the number of steps for random walks) for the pol dataset.  \textbf{B}. Elbow method for k-means for the pol dataset. \textbf{C}. accuracy comparison between clustering and random walk approaches}
        \label{fig:S7}
    \end{figure}

\vspace{\baselineskip}

\clearpage

\begin{figure}[hbt!]
        \centering
         \includegraphics[width=14cm]{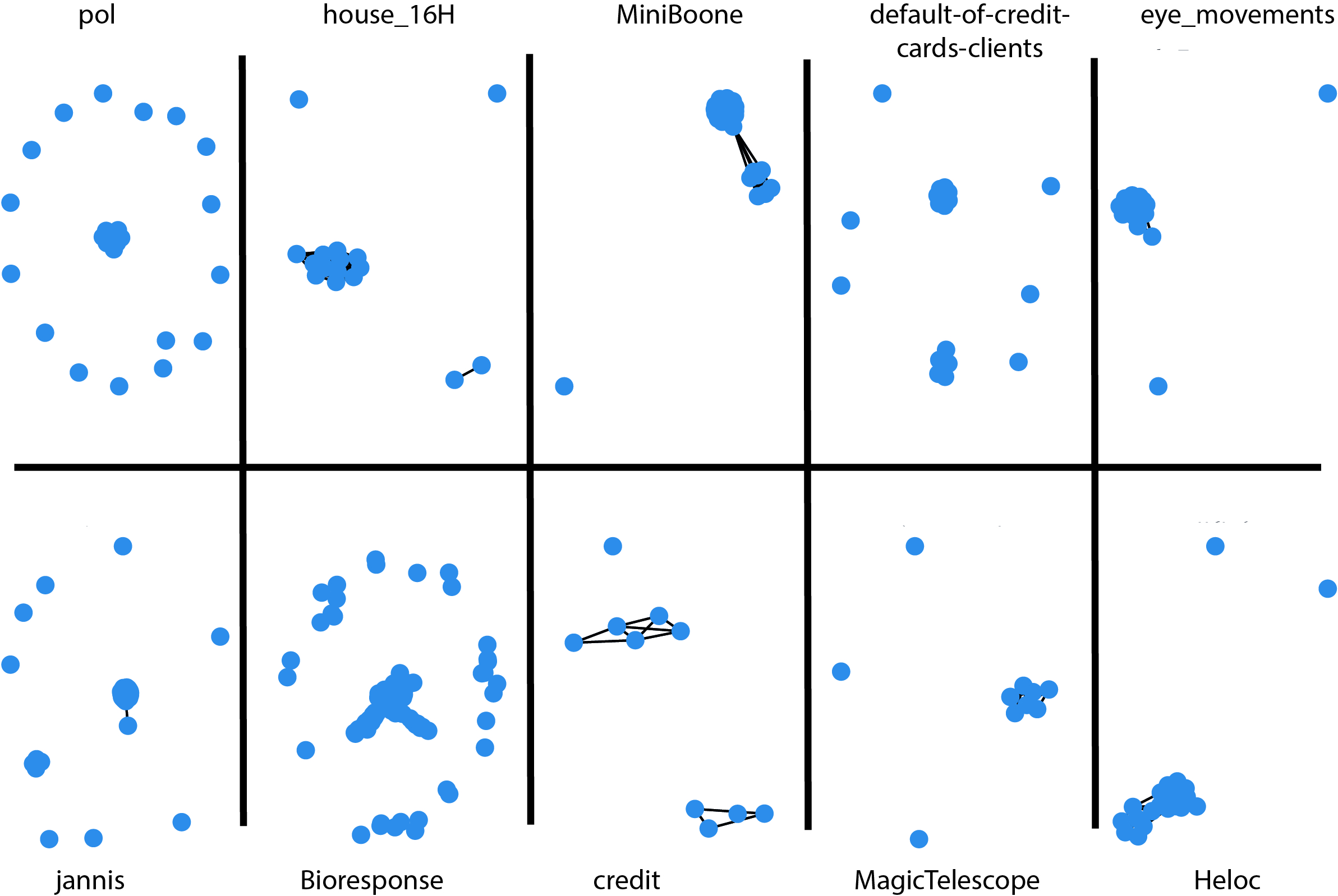}
        \caption{\textbf{Fig. S8} Example of feature graph obtained for each tabular dataset. Each graph represents a dataset obtained using our proposed algorithm. As described above, each graph is obtained using the cosine similarity as a similarity measure (we add an edge if the cosine similarity is higher than 0.5 or fewer than 0.5). As we hinted before, the algorithm is generating isolated nodes.}
        \label{fig:S8}
    \end{figure}

\vspace{\baselineskip}

\clearpage

\begin{figure}[hbt!]
        \centering
         \includegraphics[width=12cm, height=13cm]{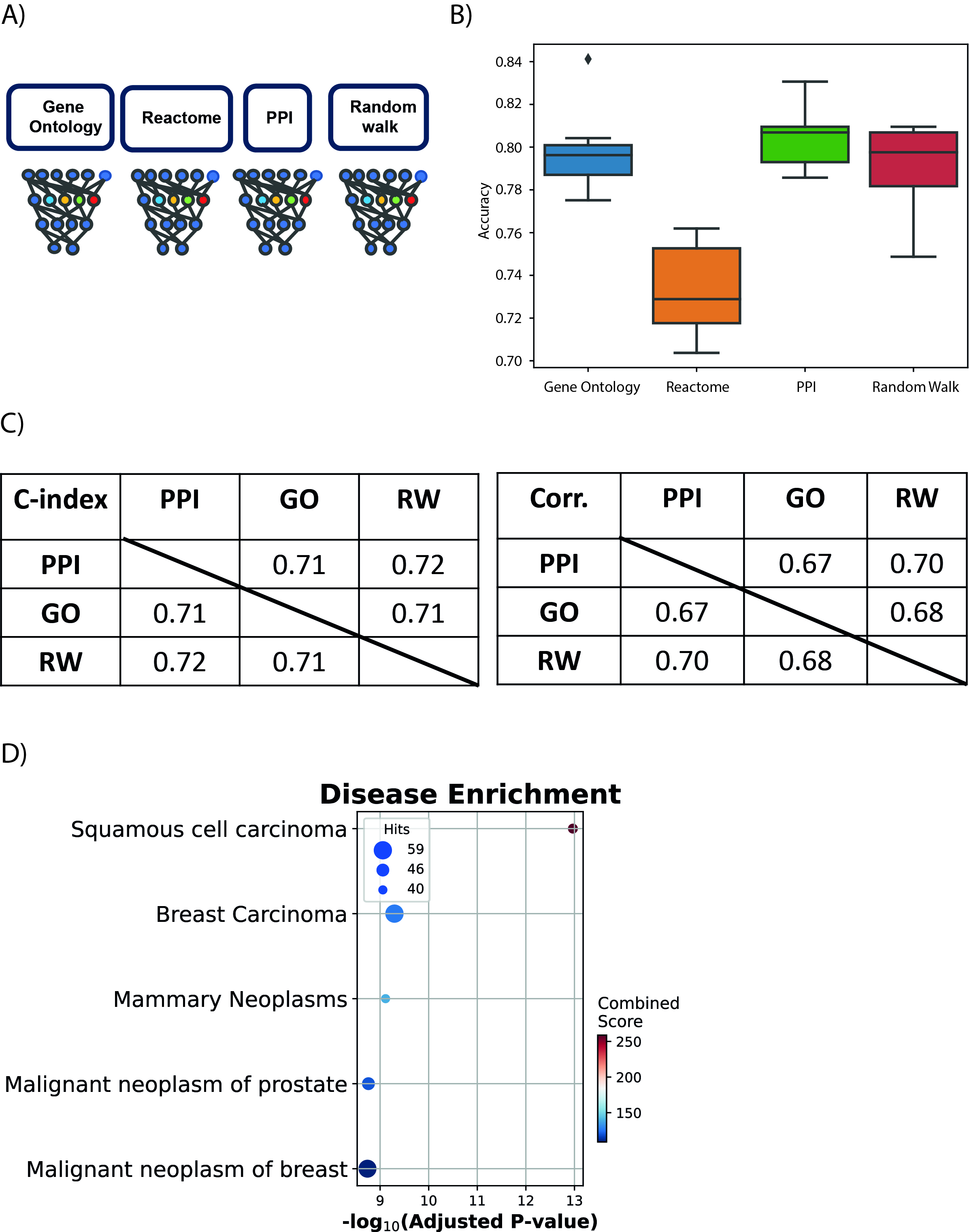}
        \caption{\textbf{Fig. S9} Unsupervised approaches have comparable performances with prior knowledge.\textbf{A}. Briefly, we used different data sources to create the architecture of the sparse net. We used Gene Ontology approach (GO), Reactome-based approach, and protein-protein interaction approach (PPI), and we compared them with our unsupervised approach: random walk-based approach (RW),  \textbf{B}. Accuracy for the sparse net for the METABRIC dataset(multi-omics multi-classification). Box plots represent the accuracy for 10-fold cross-validation. We are showing results Gene Ontology approach (GO), Reactome-based approach, random walk-based approach (RW), and protein-protein interaction approach (PPI). \textbf{C}. left:  Concordance Index for the attention weights for the genes as indicated by the GO, RW, and PPI approaches, right: Pearson correlation for the attention weights for the genes as indicated by the GO, RW, and PPI approaches. \textbf{D}. Disease enrichment for the  100 top genes according to attention importance (RW approach). }
        \label{fig:S9}
    \end{figure}

\vspace{\baselineskip}

\bibliography{sn-bibliography}

\begin{thebibliography}{10}
\expandafter\ifx\csname url\endcsname\relax
  \def\url#1{\burl{#1}}\fi
\expandafter\ifx\csname urlprefix\endcsname\relax\def\urlprefix{URL }\fi
\providecommand{\bibinfo}[2]{#2}
\providecommand{\eprint}[2][]{\url{#2}}
\providecommand{\doi}[1]{\url{https://doi.org/#1}}
\bibcommenthead

\bibitem{Borisov_2022}
\bibinfo{author}{Borisov, V.} \emph{et~al.}
\newblock \bibinfo{title}{Deep neural networks and tabular data: A survey}.
\newblock \emph{\bibinfo{journal}{{IEEE} Transactions on Neural Networks and Learning Systems}} \bibinfo{pages}{1--21} (\bibinfo{year}{2022}).

\bibitem{pgb2023}
\bibinfo{author}{Saupin, G.}
\newblock \emph{\bibinfo{title}{Practical Gradient Boosting - A deep dive into Gradient Boosting in Python}}  (\bibinfo{publisher}{AFNIL}, \bibinfo{address}{Nantes, France}, \bibinfo{year}{2023}).

\bibitem{grinsztajn2022treebased}
\bibinfo{author}{Grinsztajn, L.}, \bibinfo{author}{Oyallon, E.} \& \bibinfo{author}{Varoquaux, G.}
\newblock \bibinfo{title}{Why do tree-based models still outperform deep learning on tabular data?} (\bibinfo{year}{2022}).
\newblock \bibinfo{eprint}{{\href{https://arxiv.org/abs/2207.08815}{{arXiv:2207.08815}}}}.

\bibitem{rubachev2022revisiting}
\bibinfo{author}{Rubachev, I.}, \bibinfo{author}{Alekberov, A.}, \bibinfo{author}{Gorishniy, Y.} \& \bibinfo{author}{Babenko, A.}
\newblock \bibinfo{title}{Revisiting pretraining objectives for tabular deep learning} (\bibinfo{year}{2022}).
\newblock \bibinfo{eprint}{{\href{https://arxiv.org/abs/2207.03208}{{arXiv:2207.03208}}}}.

\bibitem{novakovsky2023obtaining}
\bibinfo{author}{Novakovsky, G.}, \bibinfo{author}{Dexter, N.}, \bibinfo{author}{Libbrecht, M.~W.}, \bibinfo{author}{Wasserman, W.~W.} \& \bibinfo{author}{Mostafavi, S.}
\newblock \bibinfo{title}{Obtaining genetics insights from deep learning via explainable artificial intelligence}.
\newblock \emph{\bibinfo{journal}{Nature Reviews Genetics}} \textbf{\bibinfo{volume}{24}}, \bibinfo{pages}{125--137} (\bibinfo{year}{2023}).

\bibitem{watson2019clinical}
\bibinfo{author}{Watson, D.~S.} \emph{et~al.}
\newblock \bibinfo{title}{Clinical applications of machine learning algorithms: beyond the black box}.
\newblock \emph{\bibinfo{journal}{Bmj}} \textbf{\bibinfo{volume}{364}} (\bibinfo{year}{2019}).

\bibitem{bayat2024pitfallsmemorizationmemorizationhurts}
\bibinfo{author}{Bayat, R.}, \bibinfo{author}{Pezeshki, M.}, \bibinfo{author}{Dohmatob, E.}, \bibinfo{author}{Lopez-Paz, D.} \& \bibinfo{author}{Vincent, P.}
\newblock \bibinfo{title}{The pitfalls of memorization: When memorization hurts generalization} (\bibinfo{year}{2024}).
\newblock \urlprefix\url{https://arxiv.org/abs/2412.07684}.
\newblock \bibinfo{eprint}{{\href{https://arxiv.org/abs/2412.07684}{{arXiv:2412.07684}}}}.

\bibitem{Roberts_2021}
\bibinfo{author}{Roberts, M.} \emph{et~al.}
\newblock \bibinfo{title}{Common pitfalls and recommendations for using machine learning to detect and prognosticate for covid-19 using chest radiographs and ct scans}.
\newblock \emph{\bibinfo{journal}{Nature Machine Intelligence}} \textbf{\bibinfo{volume}{3}}, \bibinfo{pages}{199–217} (\bibinfo{year}{2021}).
\newblock \urlprefix\url{http://dx.doi.org/10.1038/s42256-021-00307-0}.

\bibitem{somvanshi2024surveydeeptabularlearning}
\bibinfo{author}{Somvanshi, S.}, \bibinfo{author}{Das, S.}, \bibinfo{author}{Javed, S.~A.}, \bibinfo{author}{Antariksa, G.} \& \bibinfo{author}{Hossain, A.}
\newblock \bibinfo{title}{A survey on deep tabular learning} (\bibinfo{year}{2024}).
\newblock \urlprefix\url{https://arxiv.org/abs/2410.12034}.
\newblock \bibinfo{eprint}{{\href{https://arxiv.org/abs/2410.12034}{{arXiv:2410.12034}}}}.

\bibitem{bommasani2022opportunities}
\bibinfo{author}{Bommasani, R.} \emph{et~al.}
\newblock \bibinfo{title}{On the opportunities and risks of foundation models} (\bibinfo{year}{2022}).
\newblock \bibinfo{eprint}{{\href{https://arxiv.org/abs/2108.07258}{{arXiv:2108.07258}}}}.

\bibitem{frankle2019lottery}
\bibinfo{author}{Frankle, J.} \& \bibinfo{author}{Carbin, M.}
\newblock \bibinfo{title}{The lottery ticket hypothesis: Finding sparse, trainable neural networks} (\bibinfo{year}{2019}).
\newblock \bibinfo{eprint}{{\href{https://arxiv.org/abs/1803.03635}{{arXiv:1803.03635}}}}.

\bibitem{liu2023lessons}
\bibinfo{author}{Liu, S.} \& \bibinfo{author}{Wang, Z.}
\newblock \bibinfo{title}{Ten lessons we have learned in the new "sparseland": A short handbook for sparse neural network researchers} (\bibinfo{year}{2023}).
\newblock \bibinfo{eprint}{{\href{https://arxiv.org/abs/2302.02596}{{arXiv:2302.02596}}}}.

\bibitem{wysocka2023systematic}
\bibinfo{author}{Wysocka, M.}, \bibinfo{author}{Wysocki, O.}, \bibinfo{author}{Zufferey, M.}, \bibinfo{author}{Landers, D.} \& \bibinfo{author}{Freitas, A.}
\newblock \bibinfo{title}{A systematic review of biologically-informed deep learning models for cancer: fundamental trends for encoding and interpreting oncology data} (\bibinfo{year}{2023}).
\newblock \bibinfo{eprint}{{\href{https://arxiv.org/abs/2207.00812}{{arXiv:2207.00812}}}}.

\bibitem{DeloitteXAI}
\bibinfo{author}{Surkov, A.}, \bibinfo{author}{Srinivas, V.} \& \bibinfo{author}{Gregorie, J.}
\newblock \bibinfo{title}{Unleashing the power of machine learning models in banking through explainable artificial intelligence (xai)} (\bibinfo{year}{2022}).
\newblock \urlprefix\url{https://www2.deloitte.com/us/en/insights/industry/financial-services/explainable-ai-in-banking.html}.
\newblock \bibinfo{note}{Accessed on June 19, 2023}.

\bibitem{Elmarakeby_2020}
\bibinfo{author}{Elmarakeby, H.~A.} \emph{et~al.}
\newblock \bibinfo{title}{Biologically informed deep neural network for prostate cancer classification and discovery} (\bibinfo{year}{2020}).

\bibitem{reel2021using}
\bibinfo{author}{Reel, P.~S.}, \bibinfo{author}{Reel, S.}, \bibinfo{author}{Pearson, E.}, \bibinfo{author}{Trucco, E.} \& \bibinfo{author}{Jefferson, E.}
\newblock \bibinfo{title}{Using machine learning approaches for multi-omics data analysis: A review}.
\newblock \emph{\bibinfo{journal}{Biotechnology Advances}} \textbf{\bibinfo{volume}{49}}, \bibinfo{pages}{107739} (\bibinfo{year}{2021}).

\bibitem{lopez2019challenges}
\bibinfo{author}{L\'{o}pez~de Maturana, E.} \emph{et~al.}
\newblock \bibinfo{title}{Challenges in the integration of omics and non-omics data}.
\newblock \emph{\bibinfo{journal}{Genes}} \textbf{\bibinfo{volume}{10}}, \bibinfo{pages}{238} (\bibinfo{year}{2019}).

\bibitem{vahabi2022unsupervised}
\bibinfo{author}{Vahabi, N.} \& \bibinfo{author}{Michailidis, G.}
\newblock \bibinfo{title}{Unsupervised multi-omics data integration methods: a comprehensive review}.
\newblock \emph{\bibinfo{journal}{Frontiers in genetics}} \textbf{\bibinfo{volume}{13}}, \bibinfo{pages}{854752} (\bibinfo{year}{2022}).

\bibitem{attwaters2023bridging}
\bibinfo{author}{Attwaters, M.}
\newblock \bibinfo{title}{Bridging the multi-omics gap}.
\newblock \emph{\bibinfo{journal}{Nature Reviews Genetics}} \bibinfo{pages}{1--1} (\bibinfo{year}{2023}).

\bibitem{hao2018pasnet}
\bibinfo{author}{Hao, J.}, \bibinfo{author}{Kim, Y.}, \bibinfo{author}{Kim, T.-K.} \& \bibinfo{author}{Kang, M.}
\newblock \bibinfo{title}{Pasnet: pathway-associated sparse deep neural network for prognosis prediction from high-throughput data}.
\newblock \emph{\bibinfo{journal}{BMC bioinformatics}} \textbf{\bibinfo{volume}{19}}, \bibinfo{pages}{1--13} (\bibinfo{year}{2018}).

\bibitem{subramanian2005gene}
\bibinfo{author}{Subramanian, A.} \emph{et~al.}
\newblock \bibinfo{title}{Gene set enrichment analysis: a knowledge-based approach for interpreting genome-wide expression profiles}.
\newblock \emph{\bibinfo{journal}{Proceedings of the National Academy of Sciences}} \textbf{\bibinfo{volume}{102}}, \bibinfo{pages}{15545--15550} (\bibinfo{year}{2005}).

\bibitem{stuart2019integrative}
\bibinfo{author}{Stuart, T.} \& \bibinfo{author}{Satija, R.}
\newblock \bibinfo{title}{Integrative single-cell analysis}.
\newblock \emph{\bibinfo{journal}{Nature reviews genetics}} \textbf{\bibinfo{volume}{20}}, \bibinfo{pages}{257--272} (\bibinfo{year}{2019}).

\bibitem{baysoy2023technological}
\bibinfo{author}{Baysoy, A.}, \bibinfo{author}{Bai, Z.}, \bibinfo{author}{Satija, R.} \& \bibinfo{author}{Fan, R.}
\newblock \bibinfo{title}{The technological landscape and applications of single-cell multi-omics}.
\newblock \emph{\bibinfo{journal}{Nature Reviews Molecular Cell Biology}} \bibinfo{pages}{1--19} (\bibinfo{year}{2023}).

\bibitem{lim2023transitioning}
\bibinfo{author}{Lim, J.} \emph{et~al.}
\newblock \bibinfo{title}{Transitioning single-cell genomics into the clinic}.
\newblock \emph{\bibinfo{journal}{Nature Reviews Genetics}} \bibinfo{pages}{1--12} (\bibinfo{year}{2023}).

\bibitem{chattopadhyay2019gene}
\bibinfo{author}{Chattopadhyay, A.} \& \bibinfo{author}{Lu, T.-P.}
\newblock \bibinfo{title}{Gene-gene interaction: the curse of dimensionality}.
\newblock \emph{\bibinfo{journal}{Annals of translational medicine}} \textbf{\bibinfo{volume}{7}} (\bibinfo{year}{2019}).

\bibitem{george2014survival}
\bibinfo{author}{George, B.}, \bibinfo{author}{Seals, S.} \& \bibinfo{author}{Aban, I.}
\newblock \bibinfo{title}{Survival analysis and regression models}.
\newblock \emph{\bibinfo{journal}{Journal of nuclear cardiology}} \textbf{\bibinfo{volume}{21}}, \bibinfo{pages}{686--694} (\bibinfo{year}{2014}).

\bibitem{wang2019machine}
\bibinfo{author}{Wang, P.}, \bibinfo{author}{Li, Y.} \& \bibinfo{author}{Reddy, C.~K.}
\newblock \bibinfo{title}{Machine learning for survival analysis: A survey}.
\newblock \emph{\bibinfo{journal}{ACM Computing Surveys (CSUR)}} \textbf{\bibinfo{volume}{51}}, \bibinfo{pages}{1--36} (\bibinfo{year}{2019}).

\bibitem{sarkar2017analysis}
\bibinfo{author}{Sarkar, K.}, \bibinfo{author}{Chowdhury, R.} \& \bibinfo{author}{Dasgupta, A.}
\newblock \bibinfo{title}{Analysis of survival data: challenges and algorithm-based model selection}.
\newblock \emph{\bibinfo{journal}{Journal of Clinical and Diagnostic Research: JCDR}} \textbf{\bibinfo{volume}{11}}, \bibinfo{pages}{LC14} (\bibinfo{year}{2017}).

\bibitem{cox1972regression}
\bibinfo{author}{Cox, D.~R.}
\newblock \bibinfo{title}{Regression models and life-tables}.
\newblock \emph{\bibinfo{journal}{Journal of the Royal Statistical Society: Series B (Methodological)}} \textbf{\bibinfo{volume}{34}}, \bibinfo{pages}{187--202} (\bibinfo{year}{1972}).

\bibitem{mariani1997prognostic}
\bibinfo{author}{Mariani, L.} \emph{et~al.}
\newblock \bibinfo{title}{Prognostic factors for metachronous contralateral breast cancer: a comparison of the linear cox regression model and its artificial neural network extension}.
\newblock \emph{\bibinfo{journal}{Breast cancer research and treatment}} \textbf{\bibinfo{volume}{44}}, \bibinfo{pages}{167--178} (\bibinfo{year}{1997}).

\bibitem{sciki_surv}
\bibinfo{author}{P{\"o}lsterl, S.}
\newblock \bibinfo{title}{scikit-survival: A library for time-to-event analysis built on top of scikit-learn}.
\newblock \emph{\bibinfo{journal}{Journal of Machine Learning Research}} \textbf{\bibinfo{volume}{21}}, \bibinfo{pages}{1--6} (\bibinfo{year}{2020}).
\newblock \urlprefix\url{http://jmlr.org/papers/v21/20-729.html}.

\bibitem{kadra2021welltuned}
\bibinfo{author}{Kadra, A.}, \bibinfo{author}{Lindauer, M.}, \bibinfo{author}{Hutter, F.} \& \bibinfo{author}{Grabocka, J.}
\newblock \bibinfo{title}{Well-tuned simple nets excel on tabular datasets} (\bibinfo{year}{2021}).
\newblock \bibinfo{eprint}{{\href{https://arxiv.org/abs/2106.11189}{{arXiv:2106.11189}}}}.

\bibitem{shwartzziv2021tabular}
\bibinfo{author}{Shwartz-Ziv, R.} \& \bibinfo{author}{Armon, A.}
\newblock \bibinfo{title}{Tabular data: Deep learning is not all you need} (\bibinfo{year}{2021}).
\newblock \bibinfo{eprint}{{\href{https://arxiv.org/abs/2106.03253}{{arXiv:2106.03253}}}}.

\bibitem{ulmer2020trust}
\bibinfo{author}{Ulmer, D.}, \bibinfo{author}{Meijerink, L.} \& \bibinfo{author}{Cinà, G.}
\newblock \bibinfo{title}{Trust issues: Uncertainty estimation does not enable reliable ood detection on medical tabular data} (\bibinfo{year}{2020}).
\newblock \bibinfo{eprint}{{\href{https://arxiv.org/abs/2011.03274}{{arXiv:2011.03274}}}}.

\bibitem{clements2020sequential}
\bibinfo{author}{Clements, J.~M.}, \bibinfo{author}{Xu, D.}, \bibinfo{author}{Yousefi, N.} \& \bibinfo{author}{Efimov, D.}
\newblock \bibinfo{title}{Sequential deep learning for credit risk monitoring with tabular financial data} (\bibinfo{year}{2020}).
\newblock \bibinfo{eprint}{{\href{https://arxiv.org/abs/2012.15330}{{arXiv:2012.15330}}}}.

\bibitem{ahmed2017survey}
\bibinfo{author}{Ahmed, M.}, \bibinfo{author}{Afzal, H.}, \bibinfo{author}{Majeed, A.} \& \bibinfo{author}{Khan, B.}
\newblock \bibinfo{title}{A survey of evolution in predictive models and impacting factors in customer churn}.
\newblock \emph{\bibinfo{journal}{Advances in Data Science and Adaptive Analysis}} \textbf{\bibinfo{volume}{9}}, \bibinfo{pages}{1750007} (\bibinfo{year}{2017}).

\bibitem{buczak2015survey}
\bibinfo{author}{Buczak, A.~L.} \& \bibinfo{author}{Guven, E.}
\newblock \bibinfo{title}{A survey of data mining and machine learning methods for cyber security intrusion detection}.
\newblock \emph{\bibinfo{journal}{IEEE Communications surveys \& tutorials}} \textbf{\bibinfo{volume}{18}}, \bibinfo{pages}{1153--1176} (\bibinfo{year}{2015}).

\bibitem{urban2021deep}
\bibinfo{author}{Urban, C.~J.} \& \bibinfo{author}{Gates, K.~M.}
\newblock \bibinfo{title}{Deep learning: A primer for psychologists.}
\newblock \emph{\bibinfo{journal}{Psychological Methods}} \textbf{\bibinfo{volume}{26}}, \bibinfo{pages}{743} (\bibinfo{year}{2021}).

\bibitem{bahdanau2016neural}
\bibinfo{author}{Bahdanau, D.}, \bibinfo{author}{Cho, K.} \& \bibinfo{author}{Bengio, Y.}
\newblock \bibinfo{title}{Neural machine translation by jointly learning to align and translate} (\bibinfo{year}{2016}).
\newblock \bibinfo{eprint}{{\href{https://arxiv.org/abs/1409.0473}{{arXiv:1409.0473}}}}.

\bibitem{luong2015effective}
\bibinfo{author}{Luong, M.-T.}, \bibinfo{author}{Pham, H.} \& \bibinfo{author}{Manning, C.~D.}
\newblock \bibinfo{title}{Effective approaches to attention-based neural machine translation} (\bibinfo{year}{2015}).
\newblock \bibinfo{eprint}{{\href{https://arxiv.org/abs/1508.04025}{{arXiv:1508.04025}}}}.

\bibitem{mnih2014recurrent}
\bibinfo{author}{Mnih, V.}, \bibinfo{author}{Heess, N.}, \bibinfo{author}{Graves, A.} \& \bibinfo{author}{Kavukcuoglu, K.}
\newblock \bibinfo{title}{Recurrent models of visual attention} (\bibinfo{year}{2014}).
\newblock \bibinfo{eprint}{{\href{https://arxiv.org/abs/1406.6247}{{arXiv:1406.6247}}}}.

\bibitem{vaswani2017attention}
\bibinfo{author}{Vaswani, A.} \emph{et~al.}
\newblock \bibinfo{title}{Attention is all you need} (\bibinfo{year}{2017}).
\newblock \bibinfo{eprint}{{\href{https://arxiv.org/abs/1706.03762}{{arXiv:1706.03762}}}}.

\bibitem{kipf2017semisupervised}
\bibinfo{author}{Kipf, T.~N.} \& \bibinfo{author}{Welling, M.}
\newblock \bibinfo{title}{Semi-supervised classification with graph convolutional networks} (\bibinfo{year}{2017}).
\newblock \bibinfo{eprint}{{\href{https://arxiv.org/abs/1609.02907}{{arXiv:1609.02907}}}}.

\bibitem{Guyon2003DesignOE}
\bibinfo{author}{Guyon, I.} \& \bibinfo{author}{Elisseeff, A.}
\newblock \bibinfo{title}{An introduction to variable and feature selection}.
\newblock \emph{\bibinfo{journal}{Journal of Machine Learning Research}} \textbf{\bibinfo{volume}{3}}, \bibinfo{pages}{1157--1182} (\bibinfo{year}{2003}).
\newblock \urlprefix\url{http://www.jmlr.org/papers/v3/guyon03a.html}.

\bibitem{NIPS2017_8a20a862}
\bibinfo{author}{Lundberg, S.~M.} \& \bibinfo{author}{Lee, S.-I.}
\newblock \bibinfo{editor}{Guyon, I.} \emph{et~al.} (eds) \emph{\bibinfo{title}{A unified approach to interpreting model predictions}}.
\newblock (eds \bibinfo{editor}{Guyon, I.} \emph{et~al.}) \emph{\bibinfo{booktitle}{Advances in Neural Information Processing Systems}}, Vol.~\bibinfo{volume}{30} (\bibinfo{publisher}{Curran Associates, Inc.}, \bibinfo{year}{2017}).
\newblock \urlprefix\url{https://proceedings.neurips.cc/paper_files/paper/2017/file/8a20a8621978632d76c43dfd28b67767-Paper.pdf}.

\bibitem{tomsett2019sanity}
\bibinfo{author}{Tomsett, R.}, \bibinfo{author}{Harborne, D.}, \bibinfo{author}{Chakraborty, S.}, \bibinfo{author}{Gurram, P.} \& \bibinfo{author}{Preece, A.}
\newblock \bibinfo{title}{Sanity checks for saliency metrics} (\bibinfo{year}{2019}).
\newblock \bibinfo{eprint}{{\href{https://arxiv.org/abs/1912.01451}{{arXiv:1912.01451}}}}.

\bibitem{curtis2012genomic}
\bibinfo{author}{Curtis, C.} \emph{et~al.}
\newblock \bibinfo{title}{The genomic and transcriptomic architecture of 2,000 breast tumours reveals novel subgroups}.
\newblock \emph{\bibinfo{journal}{Nature}} \textbf{\bibinfo{volume}{486}}, \bibinfo{pages}{346--352} (\bibinfo{year}{2012}).

\bibitem{cancer2013cancer}
\bibinfo{author}{Network, T. C. G. A.~R.} \emph{et~al.}
\newblock \bibinfo{title}{The cancer genome atlas pan-cancer analysis project}.
\newblock \emph{\bibinfo{journal}{Nature genetics}} \textbf{\bibinfo{volume}{45}}, \bibinfo{pages}{1113--1120} (\bibinfo{year}{2013}).

\bibitem{Ashburner2000}
\bibinfo{author}{Ashburner, M.} \emph{et~al.}
\newblock \bibinfo{title}{Gene ontology: tool for the unification of biology}.
\newblock \emph{\bibinfo{journal}{Nature Genetics}} \textbf{\bibinfo{volume}{25}}, \bibinfo{pages}{25--29} (\bibinfo{year}{2000}).
\newblock \urlprefix\url{https://doi.org/10.1038/75556}.

\bibitem{wolf2018scanpy}
\bibinfo{author}{Wolf, F.~A.}, \bibinfo{author}{Angerer, P.} \& \bibinfo{author}{Theis, F.~J.}
\newblock \bibinfo{title}{{SCANPY}: Large-scale single-cell gene expression data analysis}.
\newblock \emph{\bibinfo{journal}{Genome Biology}} \textbf{\bibinfo{volume}{19}}, \bibinfo{pages}{1--5} (\bibinfo{year}{2018}).

\bibitem{yang2022fastcph}
\bibinfo{author}{Yang, X.} \emph{et~al.}
\newblock \bibinfo{title}{Fastcph: Efficient survival analysis for neural networks} (\bibinfo{year}{2022}).
\newblock \bibinfo{eprint}{{\href{https://arxiv.org/abs/2208.09793}{{arXiv:2208.09793}}}}.

\bibitem{grover2016node2vec}
\bibinfo{author}{Grover, A.} \& \bibinfo{author}{Leskovec, J.}
\newblock \bibinfo{title}{node2vec: Scalable feature learning for networks} (\bibinfo{year}{2016}).
\newblock \bibinfo{eprint}{{\href{https://arxiv.org/abs/1607.00653}{{arXiv:1607.00653}}}}.

\end{thebibliography}

\end{document}